\documentclass{article}
\usepackage{ijcai26}

\usepackage{times}
\usepackage{soul}
\usepackage{url}
\usepackage[hidelinks]{hyperref}
\usepackage[utf8]{inputenc}
\usepackage[small]{caption}
\usepackage{graphicx}
\usepackage{amsmath}
\usepackage{amsthm}
\usepackage{booktabs}
\usepackage{algorithm}
\usepackage[switch]{lineno}
\usepackage[subpreambles=true]{standalone}
\usepackage{xcolor}
\usepackage{amssymb}
\usepackage{lipsum}
\usepackage[inline]{enumitem}

\usepackage[noend]{algpseudocode}
\usepackage{nicefrac}
\usepackage{esvect}

\algblockdefx[WHEN]{When}{EndWhen}[1]{\textbf{when} #1 \textbf{do}}{}
\algtext*{EndWhen}

\usepackage{multirow}
\usepackage{subcaption}

\usepackage{mystyle}
\usepackage{macros}

\DeclareMathOperator*{\argmax}{argmax}

\newcommand{\citet}[2][]{\citeauthor[#1]{#2}\shortcite{#2}}

\title{Online Goal Recognition using Path Signature and Dynamic Time Warping}

\author{
Douglas Tesch$^{1}$\and
Nathan Gavenski$^{2}$\and
Leonardo Rosa Amado$^3$\and
Odinaldo Rodrigues$^2$\And\break
Felipe Meneguzzi$^{1,3}$\\
\affiliations
$^1$Pontifícia Universidade Católica do Rio Grande do Sul\\
$^2$King's College London\\
$^3$University of Aberdeen\\
\emails
douglas.tesch@edu.pucrs.br,
\{nathan.schneider\_gavenski, odinaldo.rodrigues\}@kcl.ac.uk,
\{leonardo.amado, felipe.meneguzzi\}@abdn.ac.uk
}

\begin{document}

\maketitle

\begin{abstract}
Online goal recognition in continuous domains poses two central challenges: efficiently encoding large trajectories and effectively comparing them.
Recent work addresses these challenges by using custom state-space representations and metrics to compare observations against hypotheses.
However, these approaches often overlook well-established encoding techniques used in other domains that offer substantial advantages.
This paper introduces a novel method for online goal recognition that leverages path signatures, a compact, expressive representation of rough path theory that efficiently captures key semantic features of trajectories, enabling more meaningful comparisons between them.
Experiments show that our method consistently outperforms the state of the art in predictive accuracy and online planning efficiency, while remaining competitive offline.
\end{abstract}


\section{Introduction}
\label{sec:introduction}

Goal recognition aims to infer the goal of an agent solely based on sparse observations along with knowledge of the environment~\cite{sukthankar2014plan}.
While research on goal recognition focuses on environments with discrete representations~\cite{Meneguzzi2021}, \textit{continuous online goal recognition}~\cite{KaminkaVeredAgmon2018} aims to support real-time inference in continuous domains while maintaining reliable goal inference.
Such approaches often compute the probability of each possible goal by building complete plans that conform to observations as they become available. 
However, these methods require multiple planner calls that scale linearly with the number of goals and observations~\cite{kaminka2018plan}, resulting in substantial computational overhead.
This overhead is impractical in applications that require fast results.
Recent approaches focus on reducing the computational cost.
\citeauthor{vered2017heuristic}~\shortcite{vered2017heuristic} used a geometric heuristic to cut unfeasible goals, whereas \citeauthor{tesch2023online}~\shortcite{tesch2023online} used a light approximation of the complete trajectories to avoid real-time (online) plan computation.

The current trend in goal recognition research in continuous domains focuses on inferring goals directly from the environment's state-space~\cite{kaminka2018plan,fitzpatrick2021behaviour,tesch2023online}.
These approaches often rely on raw state information (e.g., position, velocity, acceleration, and relative distances), which may not capture the underlying behavioral patterns necessary for robust goal inference.
Existing state encodings~\cite{slotine1991applied} often fail to capture the temporal structures or semantics of an agent's trajectory (information that is often critical for distinguishing among goals).
We address this gap by using \textit{path signature}~\cite{lyons1998differential} representations.
Path signatures encode key semantic characteristics of arbitrarily sized sequences of data, such as agents' trajectories, into unique, fixed-length arrays.
They yield compact representations that are invariant to reparametrization and naturally capture multi-scale dependencies, enabling more meaningful comparison between trajectories of varying lengths~\cite{gavenski2024cilo}.

Identifying trajectory differences is also a challenging task for goal recognition~\cite{tao2021comparative}.
These difficulties become especially pronounced in robotic applications, where agents are fast moving~\cite{fitzpatrick2021behaviour}.
One possible approach is to use similarity measures tailored to noise, varying path lengths, and non-Euclidean distances.
Goal recognition approaches often overlook these measures, despite their close affinity with goal recognition problems.
We overcome this challenge by pairing path signatures with \textit{Dynamic Time Warping} (DTW)~\cite{berndt1994using}. 

This paper develops \method (\abbrev), an approach that leverages path signatures to encode geometric and temporal properties of agents' trajectories.
\abbrev can be integrated with DTW to provide enhanced predictive performance in settings with noisy or incomplete observations.
Our contributions are:
\begin{enumerate*}[label=(\roman*)]
    \item an online goal recognition approach using path signatures; and
    \item an enhanced inference step using DTW that further improves predictive accuracy, albeit at an increased inference time.
\end{enumerate*}
\abbrev works in continuous and discrete domains, achieving state-of-the-art results in the former and remaining competitive in the latter.

\section{Background}
\label{sec:background}

\paragraph{Online Goal Recognition}
We adopt and generalise the goal recognition problem definition of \citet{Meneguzzi2021} to accommodate both continuous and discrete domains.
A \textit{goal recognition problem} $\grproblem$ is a tuple $\tuple{\domain, \goals, \initialstate, \observations}$ with the following elements.
The domain $\domain$ comprises the state space $\statespace$, the action space $\actions$, the observation space $\observationspace$, a transition function $\tf$, and a projection function $\of$.
States $s_t \in \statespace$ are vectors describing the domain at time $t$, and can be continuous or discrete.
Similarly, actions $a \in \actions$ are vectors that can be continuous or discrete.
A transition function $\tf$ defines how actions transform the environment and need not be Markovian, but in these cases $\tf: \statespace \times \actions \to \statespace$.
A projection function $\of: \statespace \to \observationspace$ maps an underlying state into an observation $o_t \in \observationspace$.
The sequence of observations $\observations$ then contains state information projected by $\of$.
Thus, an observation sequence is a finite sequence of state projections $\observations = \seq{\of(s_0), \cdots, \of(s_n)}$, where $s_0 = \initialstate$ and $s_n \in \goals$.
The set of goal hypotheses $\goals \subset \statespace$ enumerates the potential goals of the observed agent, with each goal hypothesis $\goal \in \goals$ being the final state after a successful plan.

A solution to $\grproblem$ is a probability distribution $\mathbb{P}_\goals$ over the goal hypotheses $\goals$ explaining the observations. 
Similarly, the initial state $\initialstate \in \statespace$ defines the starting state.
In discrete goal recognition problems, the state space $\statespace$ is finite and is often described using STRIPS-style Planning Domain Definition Language (PDDL) representations~\cite{fikes1971strips}. In contrast, in continuous goal recognition problems, the state space lies in the infinite set of real-valued vectors, i.e., \(\mathbb{R}^n\). 
To ensure feasibility within our problem formulation, we require a transformation of states in discrete (STRIPS-style) goal recognition problems into a vectorial representation.
We address this conversion in Sec.~\ref{sec:discrete_exp}.
To extend the classic definition to \textit{online} problems, one must only consider that any given sequence of observations $\observations$ may not be complete $\observations_{0:t}$ since the agent is still interacting with the environment.

Early goal recognition methods assume that agents act optimally, following least-cost paths toward their goal~\cite{ramirez2009plan}.
Under this assumption, goal hypotheses can be ranked by comparing the cost of optimal plans with observed behavior~\cite{ramirez2010probabilistic,masters2017cost}.
Specifically, \citeauthor{ramirez2010probabilistic}~\shortcite{ramirez2010probabilistic} compute goal probabilities from the difference in plan costs between plans that must explain the observations and unconstrained optimal plans reaching the same goal.
However, these plan-based approaches require repeated calls to a planner, i.e., one for each goal hypothesis and often for each new observation, making them computationally expensive.
To reduce computational costs, recent work reasons over precomputed \textit{trajectories} (state sequences from $\initialstate$ to goals) rather than invoking a planner for each new observation~\cite{vered2017heuristic,vered2018towards,masters2017cost,tesch2023online}.
The resulting path-based goal recognition problem $\vectorgrproblem = \tuple{\statespace, \goals, \initialstate, \alltrajectories, \observations}$ has an implicit (and potentially infinite) set of all possible trajectories $\alltrajectories$ from $\initialstate$ to $\goals$.
However, comparing observation sequences against trajectories remains challenging, as trajectories may differ in length, observations may be partial (or noisy), and different state representations may emphasize different structural properties.

\vspace{-.1cm}
\paragraph{Path Signatures}
Path Signatures~\cite{lyons1998differential} are fixed-length feature vectors that represent multidimensional time series (i.e., trajectories).
To compute a path signature $\pathsig_{1:n}$, where $n \in \mathbb{N}$, we take a trajectory $\trajectory$ of length belonging to $\{1,\ldots, n\}$, where each state is a vector in $\mathbb{R}^d$ with dimensions indexed by a collection of indices $i_1, \ldots, i_k \in \{1, \ldots, d\}$.
For example, in the Moving-AI~\cite{sturtevant2012benchmarks} dataset, each trajectory consists of $n$ states from one point to another, and each state has $3$ dimensions.
Thus, the recursively computed path signature $PS$ for the trajectory $\trajectory$ for any $k \geq 1$ and time $t$ ($1 \leq t \leq n$) as:
\begin{equation} \label{eq:get_signature}
    PS(\trajectory)^{i_1, \ldots, i_k}_{1:t} = \int_{1 < s \leq t} PS(\trajectory)_{1:s}^{i_1, \ldots, i_{k-1}} d\timelist_s^{i_k}.
\end{equation}
Therefore, the signature $\pathsig$ of a trajectory $\trajectory : [1, n] \rightarrow \mathbb{R}^d$ is the collection of all the iterated integrals of $\trajectory$:
\begin{equation} \label{eq:get_allsignature}
        \pathsignature{\trajectory}^k_{1:n} = \left[
            1, PS(\trajectory)^1_{1:n}, \cdots, PS(\trajectory)^{i_1, i_2, \cdots, i_k}_{1:n} 
        \right],
\end{equation}
where the first term is conventionally $1$, and $k$ defines the $k$-th level of the signature, which defines the finite collection of all terms  $PS(\trajectory)^{i_1, \cdots, i_k}_{1:n}$ for the multi-index of length $k$.
For example, when $k\!=\!d$, the last term would be $PS(\trajectory)_{1:n}^{d, d, \cdots, d}$.

Finally, path signatures are unique: no two path signatures are equal unless the trajectories that generate them are identical.
By leveraging this property, it is possible to construct a trajectory tree $\tree$ that encodes a set of all trajectories $\alltrajectories$ by storing all partial path signatures along each branch.
The value for the $j$th node of branch $i$ in the tree corresponds to the path signature of the first $j$ steps of trajectory $\trajectory^i \in \alltrajectories$, defined by $\tree_{i,j} = \pathsignature{\trajectory^i_{1:j}}$, where $1 \leq i \leq |\alltrajectories|$, and $1 \leq j \leq |\trajectory^i|$.
This property is useful in our setting because it guarantees a one-to-one correspondence between trajectory prefixes and their encoded signatures, enabling efficient and unambiguous representation of the trajectory space.
The supplement\footnote{Available at: \url{https://arxiv.org/abs/2605.07736}} explains path signatures in further detail.

\begin{figure}[h!p]
    \centering
    \begin{tikzpicture}

    \draw[step=1cm,gray!80,very thin] (-0.5,-0.75) grid (9.5, 4.75);
    \draw[thick, black] (-0.5, -0.75) rectangle (9.5, 4.75);
    
    \foreach \x in {0,...,9} {
    \node[anchor=north, gray!80] at (\x,-.75) {\Huge \x};
    }
    
    \foreach \y in {0,...,4} {
    \node[anchor=east, gray!80] at (-.5,\y) {\Huge \y};
    }
    
    \node (s1) [green, dot, label={[xshift=7pt, yshift=-3pt]above:{\Huge $s_1$}}] at (3, 2) {};
    \node (s2) [green, dot, label={[xshift=7pt, yshift=-3pt]above:{\Huge $s_2$}}] at (1,4) {};
    \node (s3) [green, dot, label={[xshift=7pt, yshift=-3pt]above:{\Huge $s_3$}}] at (7,3) {};
    
    \node (s'1) [seq_blue, dot, label={[xshift=15pt, yshift=19pt]below:{\Huge $s'_1$}}] at (1, 0) {};
    \node (s'2) [seq_blue, dot, label={[xshift=-5pt, yshift=0pt]above:{\Huge $s'_2$}}] at (0, 2) {};
    \node (s'3) [seq_blue, dot, label={[xshift=-10pt, yshift=0pt]above:{\Huge $s'_3$}}] at (6, 0) {};
    \node (s'4) [seq_blue, dot, label={[xshift=1pt, yshift=5pt]below:{\Huge $s'_4$}}] at (9, 3) {};
    
    \draw[edge, green] (s1) -- (s2);
    \draw[edge, green] (s2) -- (s3);

    \draw[edge, seq_blue] (s'1) -- (s'2);
    \draw[edge, seq_blue] (s'2) -- (s'3);
    \draw[edge, seq_blue] (s'3) -- (s'4);

    \draw[dedge, orange] (s1) -- (s'1);
    \draw[dedge, orange] (s2) -- (s'2);
    \draw[dedge, orange] (s3) -- (s'3);
    \draw[dedge, orange] (s3) -- (s'4);

\end{tikzpicture}
    \hspace{.5cm}

\renewcommand{\arraystretch}{1.2} 

\begin{tabular}{|c|c|c|c|c|}
\multicolumn{5}{l}{\textbf{\hspace{-10pt} Squared Distances}} \\
\hline
 & $s'_1$ & $s'_2$ & $s'_3$ & $s'_4$ \\
\hline
$s_3$ & 45 & 50 & \cellcolor{cb_orange}10 & \cellcolor{cb_orange}4 \\
\hline
$s_2$ & 16 & \cellcolor{cb_orange}5 & 41 & 65 \\
\hline
$s_1$ & \cellcolor{cb_orange}8 & 9 & 13 & 37 \\
\hline
\end{tabular}

    \caption{Dynamic Time Warping example.}
    \label{fig:dtw_example}
\end{figure}
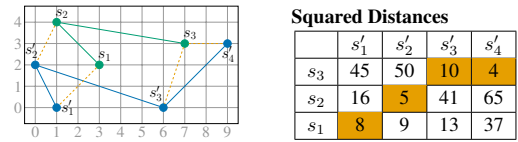

\paragraph{Dynamic Time Warping}
\cite{berndt1994using} aligns two trajectories by stretching or compressing them in the time dimension to find the best match, i.e., the timestamps with the minimum distance between the trajectories. 
\citeauthor{berndt1994using} provide a formal definition of DTW in their work: given two trajectories $\trajectory = [s_1, s_2, \cdots, s_n]$ and $\trajectory' = [s'_1, s'_2, \cdots, s'_m]$, the algorithm creates a cost matrix $n \times m$ between each pair of sample trajectories $s_i$ and $s'_j$. 
Then, with all matrix costs computed, the algorithm searches for a minimum-cost path that links the lower-left and upper-right corner elements.
Fig.~\ref{fig:dtw_example} shows the cost matrix between trajectories $\trajectory$ and $\trajectory'$ considering the squared Euclidean distance ($||s_i-s'_j||^2$).
Note that the orange cells represent the sequence with the minimum cost ($27$).
The output of the DTW algorithm is a vector combining the indices of the elements that best align the sequences.
Thus, the output is $\alignindex = \left[\left(1,1\right);\left(2,2\right);\left(3,3\right);\left(3,4\right)\right]$, knows as warping path.
\section{\method}
\label{sec:online_recognition_with_path_signature}

\abbrev has two key innovations:
\begin{enumerate*}[label=(\roman*)]
    \item tree structures from set of trajectories $\topktrajectoriesnog$ based on path signatures, further improved by applying \textit{merging} and \textit{pruning} operations to identify relevant states across trajectories; and
    \item integrating DTW as a similarity measure to compare path signatures even when they are not temporally aligned.
\end{enumerate*}
Alg.~\ref{alg:vec_inferencePS_DTW} provides an overview of \abbrev{}+DTW. 
Importantly, DTW is optional and can be bypassed when computational efficiency is critical, such as in real-time applications.
\begin{algorithm}[b!tp] 
    \footnotesize   
    \caption{Goal Recognition with PS and DTW}\label{alg:vec_inferencePS_DTW}
    \begin{algorithmic}[1]
    \Require $\domain, \goals, \initialstate$ from goal recognition problem
    \Require $\topksize$ number of top trajectories
    \Require $\mergethreshold, \prunethreshold, k$ to build trajectory tree 

    \smallskip
    \Statex \textbf{\method } \Comment{Offline}
    \State $\topktrajectoriesnog \gets [ \topktrajectories \mapsto \Call{SampleKTraj}{\domain, \initialstate, \goal, \topksize} \mid \goal \in \goals]$ \label{DTW:function_traj}
    \State $\tree \gets \Call{trajectoryTree}{\topktrajectoriesnog, \mergethreshold, \prunethreshold, k}$ \Comment{Algorithm~\ref{alg:trajectory-tree}} \label{DTW:merge_prune}

    \smallskip
    \Statex \textbf{Dynamic Time Warping} \Comment{Online}
    \State $\observations \gets [\,]$, $\signatures{\observations} \gets [\,]$
    \When{receives $o_\discretesteptime$ and $o_\discretesteptime \notin \goals$} \label{DTW:online_while}
            \State $\observations \gets \observations \cup o_t$
            \State $\signatures{\observations} \gets \signatures{\observations} \cup \pathsignature{\observations}^k_{1:t}$ \label{DTW:O_sig} \Comment{Compute observations signature}
            \State $\mathbb{P}_\goals \gets \{ g \mapsto 0 \mid g \in \goals \}$ \Comment{Initialize probabilities}
            \For{$\branch \in \tree$} \label{alg:dtw_for}
                \State $\alignindex \gets \dtw{\signatures{\observations}}{\branch}$\label{DTW:index}
                \State $g \gets \text{leaf node of } \branch$ \label{DTW:goal}
                \State $p \gets 1 - \exp\left(\nicefrac{-1}{\left\| {\pathsig_O}_t - \branch_j \right\|^2_{t, j \in \delta}}\right)$\label{DTW:line_compute}
                \State $\mathbb{P}_\goals[g] \gets \max(p, \mathbb{P}_\goals[g])$
            \EndFor \label{alg:dtw_end_for}
        \State \textbf{broadcast} $\mathbb{P}_\goals$ \label{alg:dtw_return_prop}
    \EndWhen
    \end{algorithmic}
\end{algorithm}
The offline stage begins by computing the top-$\topksize$ approximate trajectories for each goal hypothesis in the problem. 
This is analogous to recent planning-based goal and plan recognition approaches~\cite{ramirez2010probabilistic,SohrabiRiabovUdrea2016}. 
For that, we use a trajectory sampling function $\Call{SampleKTraj}{}$ that yields $\topktrajectories$ approximately optimal trajectories, given the domain $\domain$, the initial configuration $\initialstate$, a goal $\goal \in \goals$ and a pre-defined number of trajectories $\topksize$ (Alg.~\ref{alg:vec_inferencePS_DTW}, Ln.~\ref{DTW:function_traj}).\footnote{We omit $\initialstate$ for simplicity, as all trajectories share the same initial state, but keep $\goal$ since it varies across trajectory sets.}
The algorithm then uses the set $\topktrajectoriesnog$ of all top-$\topksize$ trajectories to all goals $\goals$, to build the trajectory tree $\tree$ (Alg.~\ref{alg:trajectory-tree}).
$\Call{TrajectoryTree}{}$ builds this tree by including all partial trajectories from $\topktrajectoriesnog$.
It performs both merging and pruning operations using predefined merging and pruning thresholds ($\mergethreshold$, $\prunethreshold$, explained in \ref{sec:trajectory_tree}) to reduce redundancy.

The online stage of \abbrev takes new observations $o$ from the agent at time $\discretesteptime$ and computes the partial path signature $\pathsignature{\observations}^k_{1:\discretesteptime}$ given the depth $k$ and all observations received up to that moment (Ln.~\ref{DTW:O_sig}).
\abbrev can recognize goals under partial observability, i.e., $\observations$ is not complete, so $\pathsignature{\observations}^k_{1:\discretesteptime}$ may lack the information for some (missing) time steps $\discretesteptime$. 
After computing the path signature $\signatures{\observations}$, \abbrev compares it to all branches under $\tree$ (Ln.~\ref{alg:dtw_for}-\ref{alg:dtw_end_for}).
\abbrev can use DTW to align $\observations$ with the current branch (Ln.~\ref{DTW:index}).
We leverage the FastDTW implementation~\cite{salvador2007toward}, which typically produces warping paths whose length is close to $max(\vert \signatures{\observations}\vert, \vert \branch \vert)$.
However, this implementation does not guarantee an exact length, and may exceed this value depending on the signal structure and approximation behavior.
When $\vert \alignindex \vert$ is greater than $max(\vert \signatures{\observations}\vert, \vert \branch \vert)$, multiple indices of $\branch$ may map to a single observation index of $\signatures{\observations}$. 
In such cases, we select the first occurrence of $t$ in $\alignindex$.  
For real-time inference, \abbrev can bypass the alignment considering that all $\discretesteptime$ in $\observations$ align perfectly. 
Then, it retrieves the branch's goal by looking at the leaf node (Ln.~\ref{DTW:goal}), and computes the likelihood of that goal $g$ being the goal of the agent (Ln.~\ref{DTW:line_compute}).
Finally, \abbrev returns the probability distribution over the goal hypotheses (Ln.~\ref{alg:dtw_return_prop}) from which we can extract the most likely goal up to the current time step.
For details about the complexity of our method, please refer to the supplementary material.

\subsection{Trajectory Tree, Prune and Merge}
\label{sec:trajectory_tree}

\begin{figure*}
    \centering
    \begin{subfigure}[t]{0.6\textwidth}
        \centering
        \begin{tikzpicture}[node distance=0.5cm and 1cm, font=\fontsize{16}{19.2}\selectfont]
    \node (goal0) [fill=goal0, circle, minimum size=6pt, inner sep=0pt] at (4, 3.5) {};
    \node (goal0Label) [anchor=west] at ([xshift=0.15cm]goal0.east) {Initial state $\initialstate$};

    \node (goal1) [fill=goal1, star, star points=5, star point ratio=2.25, minimum size=6pt, inner sep=0pt] at ([xshift=0.3cm]goal0Label.east) {};
    \node (goal1Label) [anchor=west] at ([xshift=0.15cm]goal1.east) {Goal $1$};

    \node (goal2) [fill=goal2, rectangle, minimum size=6pt, inner sep=0pt] at ([xshift=0.3cm]goal1Label.east) {};
    \node (goal2Label) [anchor=west] at ([xshift=0.15cm]goal2.east) {Goal $2$};

    \node (goal3) [fill=goal3, regular polygon, regular polygon sides=3, rotate=0, minimum size=6pt, inner sep=0pt] at ([xshift=0.3cm]goal2Label.east) {};
    \node (goal3Label) [anchor=west] at ([xshift=0.15cm]goal3.east) {Goal $3$};

    \node (goal4) [fill=goal4, diamond, minimum size=6pt, inner sep=0pt] at ([xshift=0.3cm]goal3Label.east) {};
    \node (goal4Label) [anchor=west] at ([xshift=0.15cm]goal4.east) {Goal $4$};

    \node (goal5) [fill=goal5, regular polygon, regular polygon sides=3, rotate=180, minimum size=6pt, inner sep=0pt] at ([xshift=0.3cm]goal4Label.east) {};
    \node (goal5Label) [anchor=west] at ([xshift=0.15cm]goal5.east) {Goal $5$};

    \node (goal6) [fill=goal6, regular polygon, regular polygon sides=3, rotate=270, minimum size=6pt, inner sep=0pt] at ([xshift=0.3cm]goal5Label.east) {};
    \node (goal6Label) [anchor=west] at ([yshift=2pt, xshift=0.15cm]goal6.east) {Goal $6$};

    \node (goal7) [fill=goal7, regular polygon, regular polygon sides=3, rotate=90, minimum size=6pt, inner sep=0pt] at ([xshift=0.3cm]goal6Label.east) {};
    \node (goal7Label) [anchor=west] at ([yshift=-2pt, xshift=0.15cm]goal7.east) {Goal $7$};
    \draw ([yshift=0.15cm, xshift=-0.1cm]goal0.north west) rectangle (goal7Label.south east);
    
    \node[tree_node] (zeroth) at (0, 0) {$0$};
    \goalsymbols{g1,g2,g3,g4,g5,g6,g7}{zeroth}

    \node[continuous_node] (first) [right=of zeroth, xshift=-.5cm] {$\cdots$};
    \node[tree_node] (204) [right=of first, xshift=-.5cm] {$204$};
    \draw [carrow] (zeroth.east) -- (first.west);
    \draw [arrow] (first.east) -- (204.west);
    \goalsymbols{g1,g2,g3,g4,g5,g6,g7}{204}

    \node[tree_node] (225) [right=of 204] {$225$};
    \node[tree_node] (217) [below=of 225] {$217$};
    \node[tree_node] (226) [above=of 225] {$226$};
    \draw [arrow] (204.east) -- ([xshift=0.375cm]204.east) |- (225.west);
    \draw [arrow] (204.east) |- ([xshift=0.375cm]204.east) |- (217.west);
    \draw [arrow] (204.east) |- ([xshift=0.375cm]204.east) |- (226.west);
    \goalsymbols{g5}{217}
    \goalsymbols{g1,g2,g3,g4}{225}
    \goalsymbols{g6,g7}{226}

    \node[tree_node] (246) [right=of 225] {$246$};
    \node[tree_node] (236) [below=of 246] {$236$};
    \draw [arrow] (225.east) -- (246.west);
    \draw [arrow] (225.east) |- ([xshift=0.66cm]225.east) |- (236.west);
    \goalsymbols{g1,g3,g4}{246}
    \goalsymbols{g2}{236}

    \node[tree_node] (262) [right=of 236] {$262$};
    \node[continuous_node] (fourth) [right=of 262, xshift=-.5cm] {$\cdots$};
    \node[tree_node] (598) [right=of fourth, xshift=-.5cm] {$598$};
    \draw [arrow] (246.east) |- ([xshift=0.66cm]246.east) |- (262.west);
    \draw [carrow] (262.east) -- (fourth.west);
    \draw [arrow] (fourth.east) -- (598.west);
    \goalsymbols{g3}{598}

    \node[tree_node] (267) [right=of 246] {$267$};
    \node[continuous_node] (fifth) [right=of 267, xshift=-.5cm] {$\cdots$};
    \node[tree_node] (342) [right=of fifth, xshift=-.5cm] {$342$};
    \draw [arrow] (246.east) -- (267.west);
    \draw [carrow] (267.east) -- (fifth.west);
    \draw [arrow] (fifth.east) -- (342.west);
    \goalsymbols{g1,g4}{267}
    \goalsymbols{g3}{262}
    \goalsymbols{g1,g4}{342}

    \node[tree_node] (354) [right=of 342] {$354$};
    \node[continuous_node] (sixth) [right=of 354, xshift=-.5cm] {$\cdots$};
    \node[tree_node] (563) [right=of sixth, xshift=-.5cm] {$563$};
    \draw [arrow] (342.east) -- (354.west);
    \draw [carrow] (354.east) -- (sixth.west);
    \draw [arrow] (sixth.east) -- (563.west);
    \goalsymbols{g4}{354}
    \goalsymbols{g4}{563}
    
    \node[tree_node] (350) [below=of 354] {$350$};
    \node[continuous_node] (seventh) [right=of 350, xshift=-.5cm] {$\cdots$};
    \node[tree_node] (562) [right=of seventh, xshift=-.5cm] {$562$};
    \draw [arrow] (342.east) |- ([xshift=.5cm]342.east) |- (350.west);
    \draw [carrow] (350.east) -- (seventh.west);
    \draw [arrow] (seventh.east) -- (562.west);
    \goalsymbols{g1}{350}
    \goalsymbols{g1}{562}

    \node[continuous_node] (eight) [right=of 226, xshift=-.5cm] {$\cdots$};
    \node[tree_node] (289) [right=of eight, xshift=-.5cm] {$289$};
    \draw [arrow] (226.east) -- (eight.west);
    \draw [arrow] (eight.east) -- (289.west);
    \goalsymbols{g6,g7}{289}

    \node[tree_node] (310) [right=of 289] {$310$};
    \draw [arrow] (289.east) -- (310.west);
    \goalsymbols{g6}{310}

    \node[tree_node] (314) [above=of 310] {$314$};
    \node[continuous_node] (ninth) [right=of 314, xshift=-.5cm] {$\cdots$};
    \node[tree_node] (455) [right=of ninth, xshift=-.5cm] {$455$};    
    \draw [arrow] (289.east) |- ([xshift=.5cm]289.east) |- (314.west);
    \draw [carrow] (314.east) -- (ninth.west);
    \draw [arrow] (ninth.east) -- (455.west);
    \goalsymbols{g7}{455}
    \goalsymbols{g7}{314}
    
    \node[continuous_node] (third) [below=of 262, xshift=-0.5cm] {$\cdots$};
    \node[tree_node] (363) [right=of third, xshift=-.5cm] {$363$};
    \draw [carrow] (236.east) |- ([xshift=0.3cm]236.east) |- (third.west);
    \draw [arrow] (third.east) -- (363.west);
    \goalsymbols{g2}{363}
    
    \node[tree_node] (504) [below=of third, xshift=-.5cm] {$504$};
    \node[continuous_node] (second) [left=of 504, xshift=.5cm] {$\cdots$};
    \draw [carrow] (217.east) |- ([xshift=0.33cm]217.east) |- (second.west);
    \draw [arrow] (second.east) -- (504.west);
    \goalsymbols{g5}{504}
    
\end{tikzpicture}
        \caption{Trajectory tree after merge and prune.}
        \label{fig:trajectory_tree_exemple_a}
    \end{subfigure}
    \hfill
    \begin{subfigure}[t]{0.39\textwidth}
        \centering
        \includegraphics[width=0.45\linewidth]{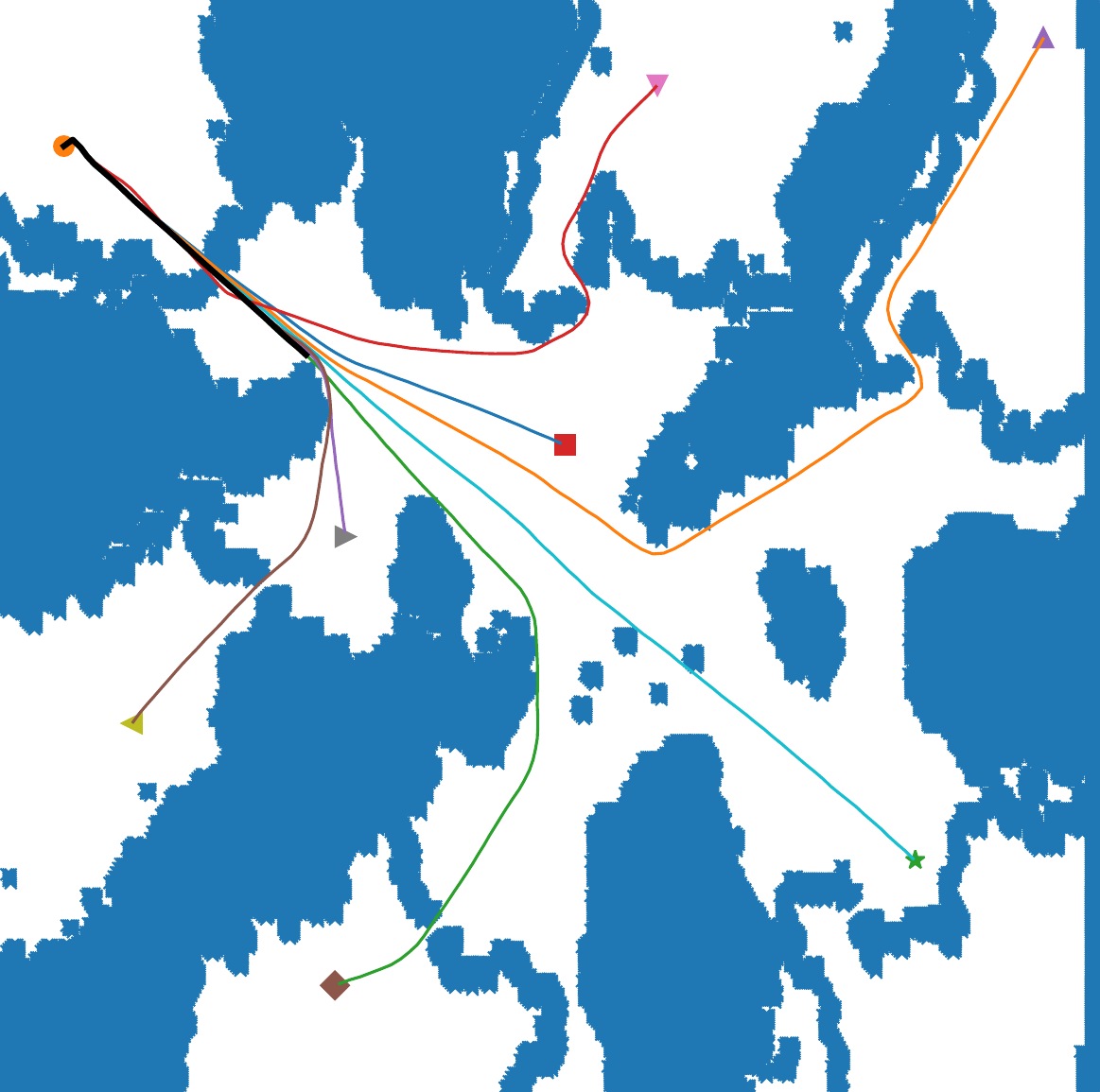}
        \caption{Example landmark (node $204$) as the black line.}
        \label{fig:trajectory_tree_exemple_b}
    \end{subfigure}
    \caption{Example of trajectory tree for the Aftershock map.}
    \label{fig:trajectory_tree_exemple}
\end{figure*}

We introduce \textit{pruning} and \textit{merging} operations on trees of path signatures.
By computing path signatures at discretely sampled intervals across multiple trajectories, we obtain a tree rooted at the initial position $\initialstate$, where nodes represent path signatures at each timestep $\discretesteptime$.
As trajectories are added, those sharing states produce identical path signatures until divergence, meaning trajectories with common prefixes share partial path signatures up to their final common point.
This enables arranging multiple trajectories in a tree where equal partial trajectories share subtrees.
A step-by-step numerical example of trajectory tree construction, including path signature computation at each node and the shared-branch structure arising from the uniqueness property, is provided in the supplementary material (Appendix~2.3--2.5).

However, path signatures differ even with slight path divergence, yielding a \textit{trajectory tree} with width between $\vert \goals \vert$ and $\vert \alltrajectories \vert$.
While this sensitivity is vital in rough path theory, it is problematic for matching semantically equivalent trajectories with minor variations: without intervention, trees may share only their root (see supplementary material for the motivation).
To address this, we perform merge and prune operations, enabling node sharing across different trajectories.
Ideally, any $\tree$ built from distinct top-$\topksize$ trajectories $\topktrajectoriesnog$ achieves width $\topksize\cdot\vert\goals\vert$, verifiable by counting leaves post-processing.

Alg.\ref{alg:trajectory-tree} (Ln.\ref{alg:merge-start}-\ref{alg:merge-end}) describes the merge operation, which compares all nodes at the same level sharing the same parent and merges those with Euclidean distance below threshold $\mergethreshold$.
Merging proceeds by selecting the rightmost node, removing it, assigning its children to the leftmost node, and computing their averaged path signature.
This operation eliminates slightly different yet closely related nodes, thus merging semantically similar trajectories.
By keeping tree levels thin, merging reduces the number of comparisons during goal recognition.
However, $\mergethreshold$ must be chosen carefully: too high a threshold can reduce tree width below $\vert \goals \vert$.
Therefore, it is vital to ensure at least as many leaves as goals.

\begin{algorithm}[b!p]
\footnotesize
\caption{Merge and Prune Operations in Trajectory Tree}
\label{alg:trajectory-tree}
\begin{algorithmic}[1]
\Require $\topktrajectoriesnog, \mergethreshold$ and $\prunethreshold$ resp. merge and prune thresholds
\Require $k$ a path signature depth
\State $\tree \gets \text{new trajectory tree with a single node with value~} 1$  
\For{each trajectory $\trajectory \in \topktrajectoriesnog$}\Comment{Trajectory Tree Creation}
    \State $n \gets \text{root of } \tree$
    \For{$\discretesteptime \gets 2 \ldots \vert \trajectory \vert$}
        \State Add new node $c$ with value $\pathsignature{\trajectory}^k_{1:\discretesteptime}$ as child of $n$; $n \gets c$
    \EndFor
\EndFor
\For{each node $n$ in $\tree$} \label{alg:merge-start}\Comment{Merge Operation}
    \For{each distinct pair $(n_z, n_j)$ of children of $n$}
        \If{$\| n^{\pathsig}_z - n^{\pathsig}_j \|^2 < \mergethreshold$}
            \State Add all of $n_j$'s children to $n_z$; Delete $n_j$
            \State $n^{\pathsig}_z \gets \nicefrac{(n^{\pathsig}_z + n^{\pathsig}_j)}{2}$
        \EndIf
    \EndFor
\EndFor \label{alg:merge-end}
\For{each node $n$ in $\tree$} \label{alg:prune-start} \Comment{Prune Operation}
    \For{each child node $n_j$ of $n_i$}
        \If{$\| n^{\pathsig} - n^{\pathsig}_j \|^2 < \prunethreshold$}
            \State Add all of $n_j$'s children to $n$; Delete $n_j$
        \EndIf
    \EndFor
\EndFor \label{alg:prune-end}
\State \Return Updated trajectory tree $\tree$
\end{algorithmic}
\end{algorithm}

The prune operation in Alg.\ref{alg:trajectory-tree}, Ln.\ref{alg:prune-start}-\ref{alg:prune-end}, occurs after all merge operations and reduces tree depth by removing nodes within a Euclidean distance $\prunethreshold$ of their parent, with the parent inheriting the pruned node's children.
Since merge operations preserve level comparisons (all siblings share timestep $\discretesteptime$), performing operations in reverse order would incorrectly compare different timesteps across trajectories.
Pruning eliminates semantically insignificant observations, e.g., an agent taking many small steps toward a goal, creating nodes that do not alter the trajectory's meaning.
Yet, pruning can reduce tree width below $\vert \goals \vert$, requiring careful selection of $\prunethreshold$.

Removing closely related nodes improves efficiency by reducing comparisons with new observations.
Goals with similar trajectories correspond to similar branches, reducing redundancy (Fig.~\ref{fig:trajectory_tree_exemple_a}).
The trajectory tree compactly represents top-$\topksize$ trajectories in continuous space, with nodes analogous to planning landmarks~\cite{ramonAIJ}.
After merging and pruning, similar goals (Fig.~\ref{fig:trajectory_tree_exemple_b}) remain in the same branch until the final bifurcation.
For example, goals $4$ and $11$ share trajectories and split last at node $342$.

\subsection{Online Goal Recognition with Path Signature}\label{sec:online_recognition_with_PS}

With a fully constructed trajectory tree $\tree$ derived from a subset of approximated trajectories $\topktrajectoriesnog \subseteq \alltrajectories$, \abbrev infers the potential goal of an observed agent (see supplementary material for further discussion).
We note that to compare the partial path signatures from $\tree$ to observations from the agent, either:
\begin{enumerate*}[label=(\roman*)]
    \item \abbrev has access to the projection function $f$, allowing it to construct a trajectory tree over observations, i.e., $f(\topktrajectoriesnog)$; or
    \item observations and states are identical, meaning the projection function is the identity, i.e., $f(s) = s$.
\end{enumerate*}

Initially, \abbrev computes the likelihood of a branch $\branch \in \tree$ being the current sequence of observations $\signatures{\observations}$:
\begin{equation}\label{eq:prob_grps}
    \probability{\branch \!\mid\! \signatures{\observations}} \!=\!
    \rho\,
    \probability{\signatures{\observations} \!\mid\! \branch}\,
    \probability{\branch \!\mid\! \goal}\probability{\goal},
\end{equation}
where $\rho$ is a normalizing term, $g$ is the goal for that branch, and the final term is a prior probability that an agent might choose each goal~\cite{ramirez2010probabilistic}. 
This might express a known preference over goals, but is often simply a uniform probability distribution.
Here, the probability $\probability{\branch \mid g}$ is always one, since each $\branch$ can only have one possible goal (leaf), and a uniform prior $\probability{\goal}$ avoids biases towards any specific, i.e., all goals are equally likely.
Since \abbrev does not use $\Call{DTW}{}$, we assume that time steps are perfectly synchronized, i.e., the observed agent's step size is equivalent to the approximated trajectories' step.
Therefore, the key part from Eq.~\ref{eq:prob_grps} is how to compute $\mathbb{P}(\signatures{\observations} \mid \branch)$ since we know all other terms. 
For this we use Eq.~\ref{eq:prob_grps_p}, which is similar to equation from Line~\ref{DTW:line_compute} from Alg.~\ref{alg:vec_inferencePS_DTW}.
\begin{equation}\label{eq:prob_grps_p}
    \mathbb{P}(\signatures{\observations}\vert \branch) =
        1 - \exp\left(\nicefrac{-1}{\left\| {\pathsig_O}_t - \branch_t \right\|^2}\right),
\end{equation}
\abbrev remains efficient in iterations of Eq~\ref{eq:prob_grps_p} over path signatures by computing the distance between ${\signatures{\observations}}_t$ and $\branch_t$ once instead of all $o \in \observations$ as in \citeauthor{fitzpatrick2021behaviour}.

Rather than averaging probabilities across branches, \abbrev uses the maximum likelihood to avoid diluting high probabilities from paths close to the observations with lower probabilities from distant ones.
This becomes important when the trajectory tree is built from a large or diverse set of trajectories (as produced by $\Call{SampleKTraj}{}$ in Alg.~\ref{alg:vec_inferencePS_DTW} Ln.~\ref{DTW:function_traj}), which results in many branches per goal--up to $\topksize \cdot \vert\goals\vert$.
In these cases, averaging probabilities across all branches that lead to the same goal can dilute high-confidence signals, especially when many branches yield low scores due to suboptimal trajectories from $\initialstate$ to $\goals$.
Since path signatures encode rich geometric and temporal structures, even beyond what is captured by the projection function $f$, their scores can vary substantially between branches.

\subsection{Dynamic Time Warping Integration} \label{sec:online_recognition_with_PS_DTW}

Assuming that the sequence of observations $\observations$ is perfectly synchronized with all trajectories in $\tree$ might not be realistic.
Therefore, \abbrev integrates DTW to solve two main related issues:
\begin{enumerate*}[label=(\roman*)]
    \item trajectories and observations that do not share the same characteristics, and
    \item potentially missing time steps from observations.
\end{enumerate*}
The first issue concerns how the synchronicity assumption in \abbrev{} can be easily violated when trajectories are sampled from agents with different characteristics, such as agents with different velocities or movement patterns from the observed agent (dynamic model error).
Since \abbrev makes no assumptions about the sampling process, it relies on the information encoded in path signatures to correctly synchronize observations and trajectories. 
However, merging and pruning also reduce the number of available steps for comparison across each trajectory.
Thus, having a function that best aligns indices between one sequence of observations and another sequence of states is crucial for improving performance.
The second issue relates to \abbrev's online step, which infers a goal only upon receiving a new observation from the agent.
Under partial observability, observations may have arbitrary gaps in time, resulting in an incomplete sequence of observations, such as $\observations = \left[o_0, o_1, o_3\right]$, where $o_2$ is missing.

We address the limitation of signature-only comparisons using Dynamic Time Warping (DTW) in Alg.~\ref{alg:vec_inferencePS_DTW} as a similarity measure to better align observations with trajectory branches.
This effectively replaces Eq.~\ref{eq:prob_grps_p} with Eq.~\ref{eq:prob_grps_p_dtw}.
\begin{equation}\label{eq:prob_grps_p_dtw}
    \mathbb{P}(\signatures{\observations}\vert \branch) \!=\! 1 - \exp\left(\nicefrac{-1}{\left\| {\pathsig_O}_t - \branch_j \right\|^2_{t, j \in \delta}}\right),
\end{equation}
where $\delta$ is the alignment path computed by DTW, consisting of index pairs $(i, j)$ that optimally align each observation index $i$ with a corresponding index $j$ in the trajectory $\branch$.
However, DTW introduces a non-negligible computational overhead.
For each $\branch \in \tree$, DTW has a time complexity of $\mathcal{O}(\vert\observations\vert \cdot \vert \branch\vert)$, leading to an overall complexity of $\mathcal{O}(\treewidth{\tree} \cdot \vert\observations\vert \cdot \treeheight{\tree})$ (Alg.~\ref{alg:vec_inferencePS_DTW}, Ln.~\ref{alg:dtw_for}--\ref{alg:dtw_end_for}).
When combined with the cost of computing path signatures—$\mathcal{O}(\vert\observations\vert \cdot d^k)$ -- the resulting inference process becomes relatively expensive for real-time applications.
We note that for partial observability settings, \abbrev{}+DTW first interpolate all observations to create an approximated sequence of observations of size $t$ ($\observations_{1:t}$).
That is, if the initial observations were $\left[\initialstate, o_2\right]$ for $t = 2$, the approximated observations would be $\left[\initialstate, \hat{o}_1, o_2\right]$, where $\hat{o}_1$ is the interpolated observation.
We provide a detailed version of Alg.~\ref{alg:vec_inferencePS_DTW} in the supplementary material.

\begin{table*}[b!tp]
\centering
\scriptsize
\begin{tabular*}{\textwidth}{l@{\extracolsep{\fill}} rrrrr @{\extracolsep{\fill}} rrrrrr}
\toprule \noalign{\vskip -2pt}
& \multicolumn{5}{c}{\textbf{(a) Continuous Domains}} & \multicolumn{6}{c}{\textbf{(b) Discrete Domains}} \\ \cmidrule{1-1} \cmidrule{2-6} \cmidrule{7-12}
Metrics & Mirr. & R+P & Vector & \abbrev & + DTW & Mirr. & Land. & GM+L & Vector & GRPS & + DTW \\ 
\cmidrule{1-1} \cmidrule{2-6} \cmidrule{7-12} \\[-10pt]
\multirow{2}{*}{ PPV (\%)} & $38.7$	& $42.1$ & $47.9$	& $50.7$ & $\textbf{53.6}$
                               & $47.3$ & $44.8$ & $48.2$ & $49.2$ & $\textbf{52.0}$ & $51.9$ \\ [-2pt]
                                                                & $\pm\, 7.0$ & $\pm\, 5.7$ & $\pm\, 9.0$ & $\pm\, 9.4$ & $\mathbf{\pm\, 8.5}$     & $\pm\, 13.6$ & $\pm\,11.2$ & $\pm\,11.5$ & $\pm\,8.9$ & $\pm\,\textbf{10.5}$ & $\pm\,10.4$ \\ [-2pt]
\cmidrule{1-1} \cmidrule{2-6} \cmidrule{7-12} \\[-10pt]
\multirow{2}{*}{ ACC (\%)} & $84.6$	& $85.4$ & $86.4$ &	$85.9$ & $\mathbf{86.7}$
                               & $81.8$ & $82.7$ & $84.7$ & $84.2$ & $\textbf{85.3}$ & $\textbf{85.3}$ \\ [-2pt]
                                                                & $\pm\, 1.8$ & $\pm\, 1.4$ & $\pm\, 2.1$ & $\pm\, 2.7$ & $\mathbf{\pm\, 2.4}$      & $\pm\,8.1$ & $\pm\,5.2$ & $\pm\,5.0$ & $\pm\,5.7$ & $\pm\,\textbf{6.0}$ & $\pm\,\textbf{6.0}$ \\[-2pt]
\cmidrule{1-1} \cmidrule{2-6} \cmidrule{7-12}\\[-10pt]
\multirow{2}{*}{ SPR}  & $1.0$ & $1.1$ & $\mathbf{1.0}$ & $\mathbf{1.0}$ & $\mathbf{1.0}$                                  & $2.5$ & $1.6$ & $\textbf{1.3}$ & $1.7$ & $1.6$ & $1.6$ \\ [-2pt]
                                                            & $\pm\, 0.0$ & $\pm\, 0.02$ & $\mathbf{\pm\, 0.0}$ & $\mathbf{\pm\, 0.0}$ & $\mathbf{\pm\, 0.0}$   & $\pm\,1.2$ & $\pm\,0.51$ & $\pm\,\textbf{0.47}$ & $\pm\,0.44$ & $\pm\,0.47$ & $\pm\,0.47$ \\ [-2pt]
\cmidrule{1-1} \cmidrule{2-6} \cmidrule{7-12}\\[-10pt]
\multirow{2}{*}{ PC}  & $49.0$ & $30.1$ & $\mathbf{7.0}$ & $\mathbf{7.0}$ & $\mathbf{7.0}$                                & $124.8$ & $\textbf{0.0}$ & $29.9$ & $8.2$ & $8.2$ & $8.2$ \\ [-2pt]
                                                            & $\pm\, 0.0$ & $\pm\, 1.6$ & $\mathbf{\pm\, 0.0}$ & $\mathbf{\pm\, 0.0}$ & $\mathbf{\pm\, 0.0}$    & $\pm\,57.4$ & $\pm\,\textbf{0}$ & $\pm\,7.9$ & $\pm\,3.9$ & $\pm\,3.9$ & $\pm\,3.9$ \\ [-2pt]
\cmidrule{1-1} \cmidrule{2-6} \cmidrule{7-12}\\[-10pt]
\multirow{2}{*}{ Online (s)}   & $1.2\text{e}{4}$ & $5.9\text{e}{3}$ & $3.9\text{e-}2$ & $\mathbf{3.0\text{e-}2}$ & $20.5$                                     & $80.4$ & $5.7\text{e-}1$ & $12.5$ & $\textbf{6.4\text{e-}2}$ & $10.1$ & $97.6$ \\ [-2pt]
                                                                    & $\pm\, 6.2\text{e}{3}$ & $\pm\, 3.2\text{e}{3}$ & $\pm\, 7.0\text{e-}4$ & $\mathbf{\pm\, 2.1\text{e-}3}$ & $\pm\, 6.6$        & $\pm\,117.6$ & $\pm\,6.1\text{e-}1$ & $\pm\,12.5$ & $\pm\,\textbf{1.3\text{e-}1}$ & $\pm\,27.4$ & $\pm\,253.8$ \\ [-2pt]
\cmidrule{1-1} \cmidrule{2-6} \cmidrule{7-12}\\[-10pt]
\multirow{2}{*}{ Offline (s)}  & $2.9\text{e}{3}$ & $2.8\text{e}{3}$ & $\mathbf{1.7\text{e}{2}}$ & $3.3\text{e}{2}$ & $2.8\text{e}{2}$              & $4.6$ & $\textbf{1.3\text{e-}1}$ & $4.8$ & $33.66$ & $60.6$ & $62.9$ \\ [-2pt]
                                                                    & $\pm\, 1.8\text{e}{3}$ & $\pm\, 1.7\text{e}{3}$ & $\mathbf{\pm\, 38.8}$ & $\pm\, 70.7$ & $\pm\, 55.7$              & $\pm\,5.3$ & $\pm\,\textbf{2.6\text{e-}1}$ & $\pm\,5.8$ & $\pm\,63.7$ & $\pm\,76.9$ & $\pm\,80.1$ \\ [-2pt]
\bottomrule
\end{tabular*}
\caption{Comparison of methods across (a) continuous and (b) discrete domains.}
\label{tab:planner_compare}
\end{table*}
\section{Experiments}
\label{sec:experiments}

We benchmark using accuracy (ACC), the spread of goals in the output (SPR), the number of planner calls (PC) required to infer each goal, Positive Predictive Value (PPV), and the runtime.
\textit{ACC} is the proportion of instances a method infers the goal correctly.
\textit{SPR} is the average number of distinct goals predicted across all instances, indicating how concentrated or diffuse the method's predictions are over the space of goal hypotheses (with a minimum value of 1).
\textit{PC} denotes the number of external planner calls required during inference, serving as a proxy for the computational overhead of planner-based methods.
\textit{PPV} measures the proportion of correctly inferred goals among all inferred predictions, i.e., $\text{PPV} = \nicefrac{TP}{TP+FP}$, where $TP$ and $FP$ denote true and false positives, respectively.
Runtime comprises two components: the \textit{offline stage}, which includes all preprocessing steps, and the \textit{online stage}, which captures the time required for goal inference per observation sequence. 

\subsection{Continuous Domains}

The experiments use the dataset from Moving-AI~\cite{sturtevant2012benchmarks}, which features a simulated yet realistic benchmark comprising $28$ scenarios from StarCraft.
We select this dataset to ensure a fair comparison between \abbrev and previous work, and elaborate on it and how we create trajectories in the supplement.
Our experiments use the trajectory sampling procedure from \citet{tesch2023online} as the $\Call{SampleKTraj}{}$ function.
We evaluate both \abbrev and \abbrev{}+DTW by varying the trajectory tree parameters, the merge threshold $\mergethreshold$, the prune threshold $\prunethreshold$, and the number of sampled trajectories per goal $\topksize$ for the sampling function.
To identify the most effective parameter combinations, we conduct a grid search across $10$ scenarios, ranging threshold values from $0$ to $2$ in increments of $0.2$, and $\topksize$ values in $\left\{ 1, 5, 10, 15 \right\}$.
The experiments use a depth of $ k=2$ for the path signature, with an ablation provided in the supplement.

\paragraph{Parameter Search}
Fig.~\ref{fig:grid_search_1} and~\ref{fig:grid_search_15} show the average PPV among the scenarios for the whole grid search space using $\topksize$ of $1$ and $15$, respectively.
This indicates that the merge variable has a limited influence on recognition process accuracy (rather than efficiency), since the merge process combines similar trajectories. 
Thus, when the number of trajectories is $\topksize=1$, there are no additional trajectories available for merging.
By contrast, Figure~\ref{fig:grid_search_15} shows the impact of the merge and prune variables when $\topksize=15$, indicating that some level of merging and pruning benefits \abbrev, while higher thresholds decrease its performance, which is to be expected.
Importantly, it shows that increasing merging has a limited impact on performance, whereas increasing pruning substantially degrades PPV.
This suggests that \abbrev is more sensitive to reductions in trajectory depth than to reductions in width.
In particular, pruning removes deeper prefixes from the trajectory tree, which are often essential for distinguishing between goals, especially when observations are partial or early in the trajectory. 
On the other hand, merging only collapses similar branches, which tends to preserve overall structure unless applied too aggressively.
As a result, while some merging can help reduce redundancy without harming performance, even moderate levels of pruning can lead to underfitting and suppressing informative variations.

\begin{figure}[h!bp]
    \centering
    \begin{subfigure}[b]{0.48\linewidth}
        \centering
        \includegraphics[width=\linewidth]{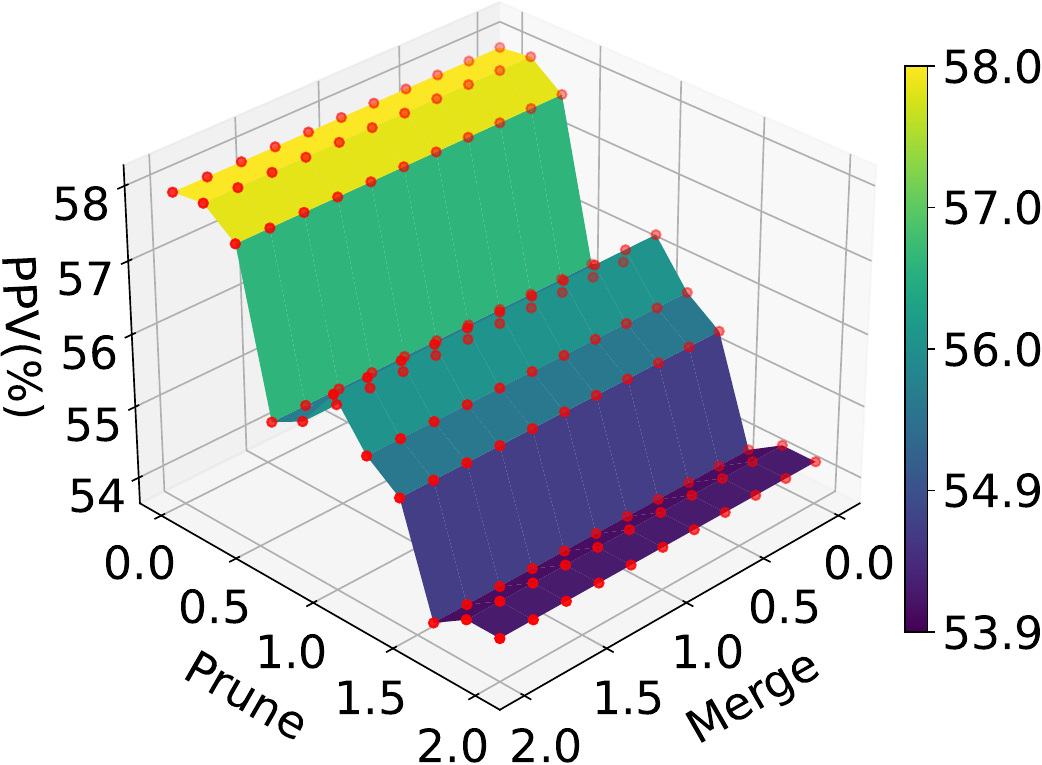}
        \caption{\abbrev with $\topksize=1$.}
        \label{fig:grid_search_1}
    \end{subfigure}
    \hfill
    \begin{subfigure}[b]{0.48\linewidth}
        \centering
        \includegraphics[width=\linewidth]{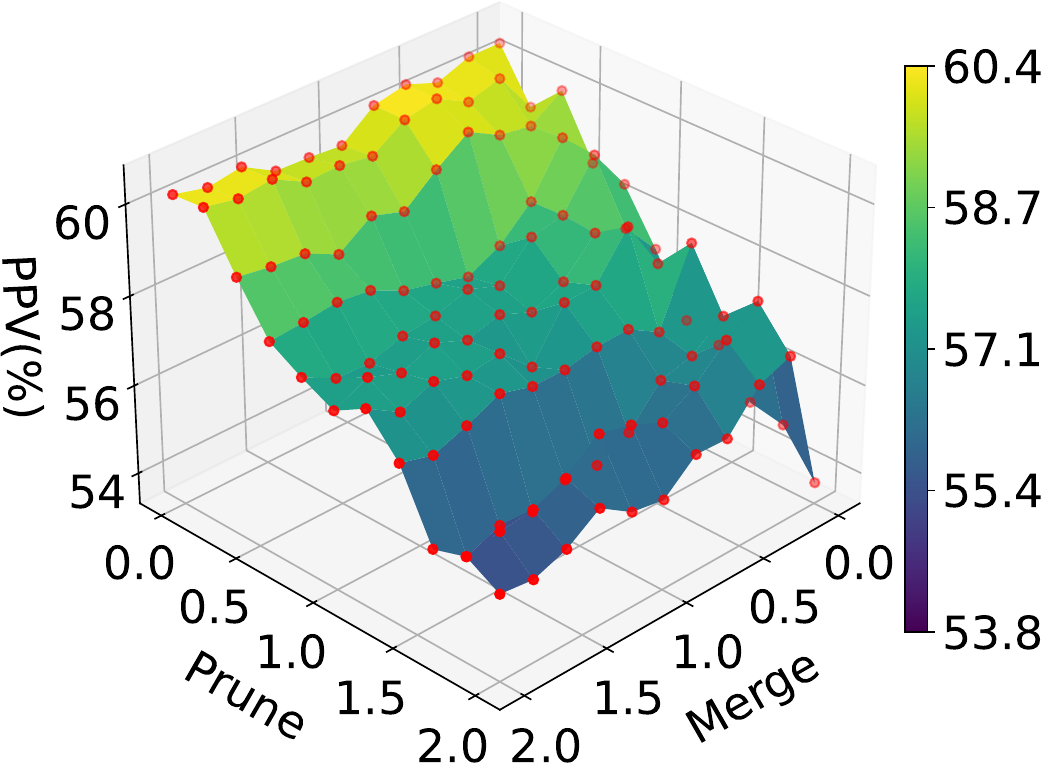}
        \caption{\abbrev with $\topksize=15$.}
        \label{fig:grid_search_15}
    \end{subfigure}
    \\[6pt]
    \begin{subfigure}[b]{0.48\linewidth}
        \centering
        \includegraphics[width=\linewidth]{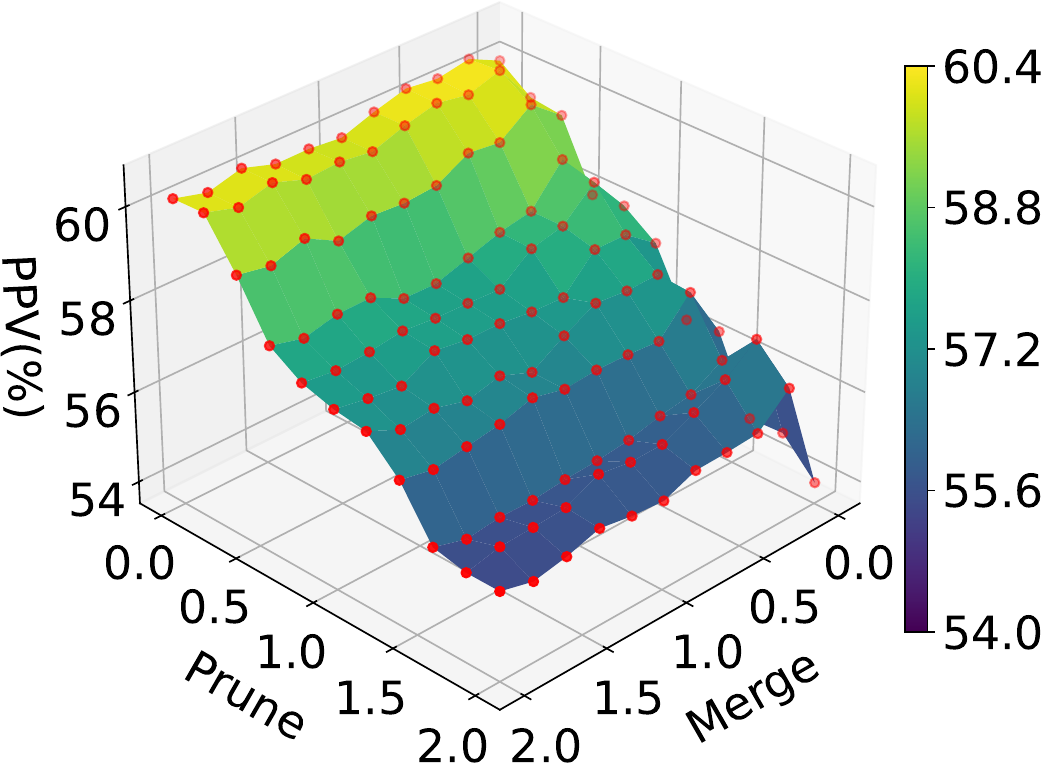}
        \caption{Average for \abbrev.}
        \label{fig:grid_all_PS}
    \end{subfigure}
    \hfill
    \begin{subfigure}[b]{0.48\linewidth}
        \centering
        \includegraphics[width=\linewidth]{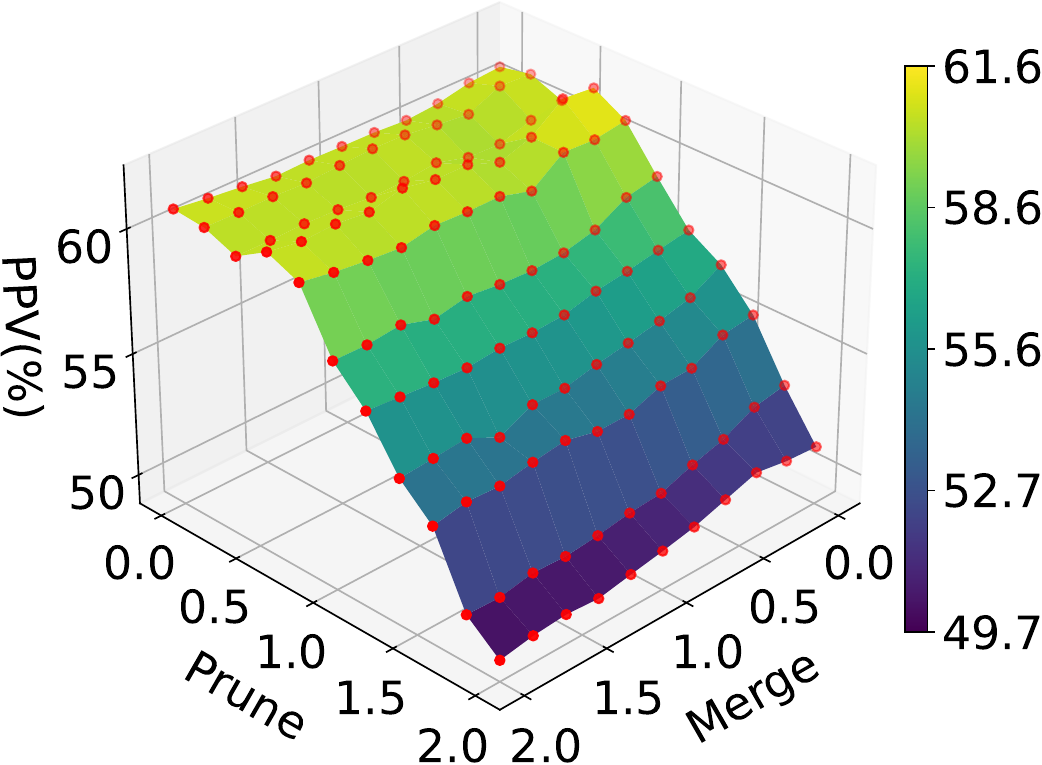}
        \caption{Average for \abbrev{}+DTW.}
        \label{fig:grid_all_DTW}
    \end{subfigure}
    \caption{Grid search results.}
    \label{fig:grid}
\end{figure}
 
Fig.~\ref{fig:grid_all_PS} and~\ref{fig:grid_all_DTW} depict the average grid search results for both inference modes for $K\!=\!1$ and $K\!=\!15$.
These results show that increasing the prune threshold harms inference quality, as measured by PPV.
However, small pruning values in the range of $0.0$ to $0.2$ offer a distinct advantage by modestly reducing tree depth without significantly harming predictive performance.
This balance is especially relevant in online inference, where shallower branches lead to faster traversal and lower computational overhead.
Thus, selecting appropriate thresholds is not only a matter of accuracy but also efficiency.
Based on these experiments, we adopt pruning and merging values of $0.2$ when using \abbrev, and $0.6$ for pruning and $0.2$ for merging when using \abbrev{}+DTW, striking a balance between preserving informative structure and maintaining inference efficiency.

\paragraph{Continuous Results}
Tab.~\ref{tab:planner_compare}a compares the results of our two inference modes with recent work:
Vector~\cite{tesch2023online}, Mirroring~\cite{kaminka2018plan}, and Recompute plus Prune (R+P)~\cite{vered2017heuristic}.
Tab.~\ref{tab:planner_compare}a shows the average values across all scenarios in the Moving-AI dataset, excluding the $10$ scenarios used for ablation. 
We report the mean of each metric (and its standard deviation) across all experiments (see supplement for details).

\abbrev{}+DTW achieves the highest PPV ($53.6\%$) and ACC ($86.7\%$) across all evaluated method, with \abbrev closely following at $50.7\%$ PPV and $85.9\%$ ACC, both surpassing the current state-of-the-art (Vector). 
This improvement in PPV is particularly significant, as it reflects the method's ability to generate correct predictions.
Unlike ACC, which is affected by class imbalance and rewards methods that often predict the majority class, PPV penalizes false positives and better reflects precision at early or ambiguous nodes, where goal recognition is most challenging. 
Notably, \abbrev achieves this improvement in PPV while also reducing Online inference time (from $3.9\text{e-}2$s to $3.0\text{e-}2$s).
Both inference modes maintain perfect prediction focus (SPR = $1.0$) and require the fewest planner calls ($\text{PC} =7.0$), indicating that they remain computationally efficient as predictive quality improves.

These results reinforce the central motivation behind GRPS: the need for more meaningful and coherent comparisons between agent behavior and candidate trajectories in goal recognition tasks.
By leveraging path signatures, \abbrev captures rich geometric and temporal dynamics while remaining robust to suboptimal behavior.
Importantly, this added expressiveness does not compromise runtime performance; \abbrev remains lightweight and fast enough for real-time applications, with an average inference time of just $30$ms.
Although \abbrev incurs slightly higher Offline times, this is primarily due to its use of Vector's sampling function for generating top-$K$ trajectories.
This component, while effective, is not intrinsic to our method. 
For example, adopting a more efficient sampling strategy or using a dataset of past trajectories could reduce the offline cost of \abbrev without affecting its inference (further discussed in supplement).
Moreover, \abbrev{}+DTW builds upon this strength by introducing dynamic temporal alignment, enabling even more accurate trajectory-observation comparisons.
While it incurs a higher online cost ($20.5$s), the resulting improvements in both PPV and ACC make it ideal for use in offline or high-precision inference settings.
Together, both modes offer a versatile framework that adapts to diverse constraints.
This dual capability demonstrates that our approach not only advances the state-of-the-art but also provides a practical and generalizable solution to goal recognition.

\begin{figure}
    \centering
    \begin{subfigure}[t]{0.9\linewidth}
        \centering
        \includegraphics[width=\linewidth]{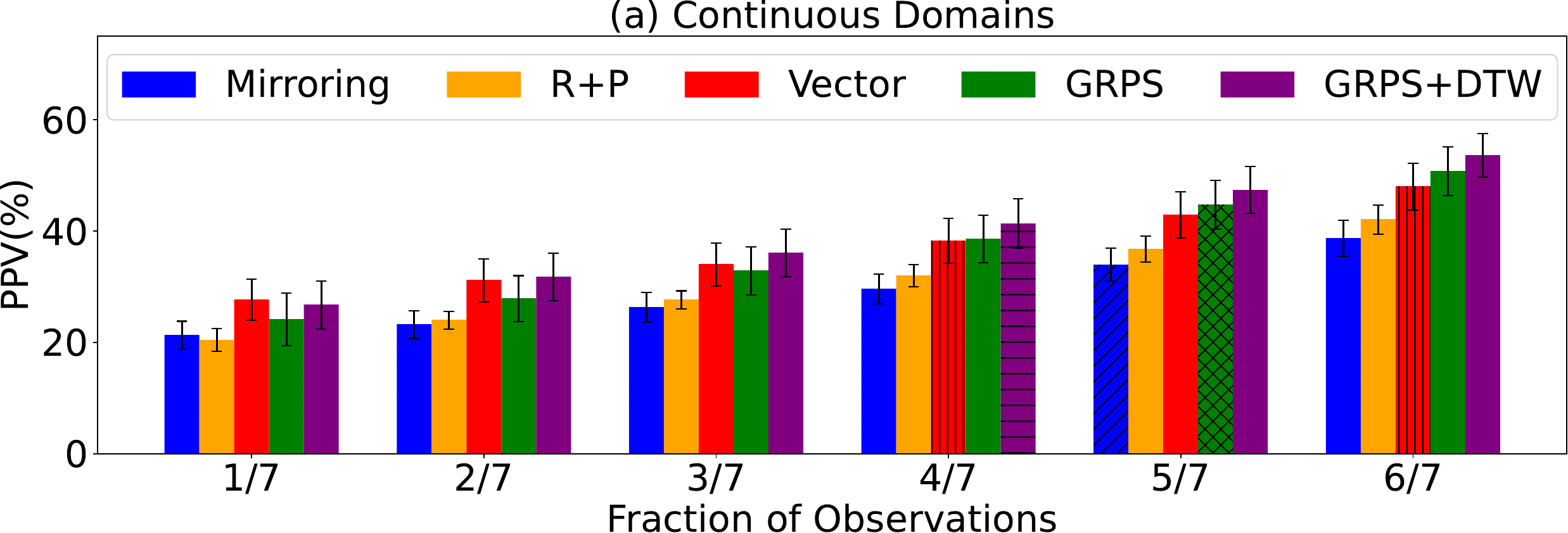}
    \end{subfigure}
    \begin{subfigure}[t]{0.9\linewidth}
        \centering
        \includegraphics[width=\linewidth]{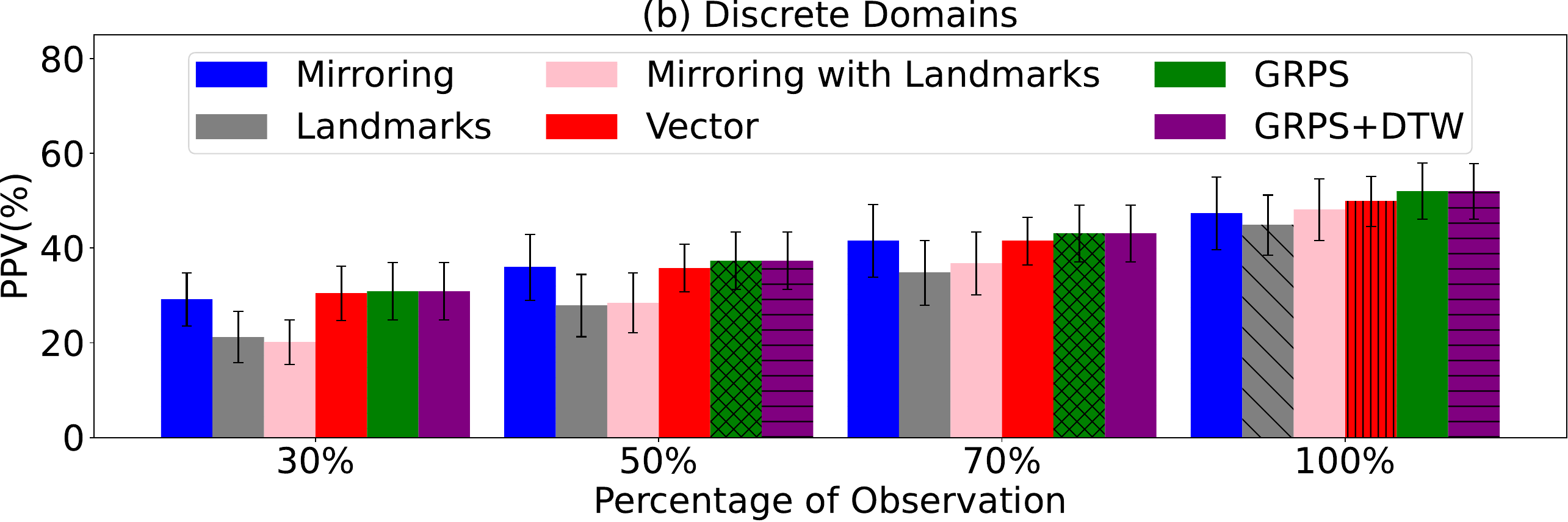}
    \end{subfigure}
    \caption{Observation fraction results for both domains.}
    \label{fig:comp_over_observations}
\end{figure}

Fig.~\ref{fig:comp_over_observations}a compares the approaches by their PPV (and its margin of error with a confidence level of $95\%$), at observation fractions of $\nicefrac{1}{7}$ to $\nicefrac{6}{7}$ of the full observation length.
\abbrev{} is competitive at initial observability stages ($\nicefrac{1}{7}$, $\nicefrac{2}{7}$) and outperforms the other methods at mid-end stages of observability.
\abbrev{}+DTW outperforms the other methods at almost all stages of observability, except at the initial stage ($\nicefrac{1}{7}$), where the available information is insufficient to provide a reliable path signature representation.

\subsection{Discrete Domains}
\label{sec:discrete_exp}

The discrete-domain experiments use the goal and plan recognition dataset~\cite{santos2021lpbased}.
We select this dataset to ensure a fair comparison with previous work. 
The dataset comprises thousands of goal-recognition problems across large, non-trivial planning instances in the STRIPS fragment of PDDL. 
It incorporates domains and problems from established benchmark datasets, including those proposed by~\citet{ramirez2009plan}. 
Tab.~\ref{tab:planner_compare}b compares our two inference methods with recent work on goal recognition:
Vector~\cite{tesch2023online}, Mirroring~\cite{kaminka2018plan}, and Goal Recognition with Landmarks and Goal Mirroring with Landmarks both from~\citet{vered2018towards}. 
We use SymK~\cite{SpeckMattmuellerNebel2020} as the sampler. 

There are two important details to consider when applying our goal recognition formulation to discrete-domain problems.
First, a direct implementation in discrete STRIPS environments is infeasible, as our formulation assumes vector- and numeric-based representations rather than symbolic descriptions. 
To overcome this limitation, we use a vectorization of the STRIPS representation~\cite{asai2018classical}, that encode states as fixed-size binary vectors over the complete grounding of predicates. 
Second, our experiments show that binary vector representations are insufficient to discriminate among trajectories.
This limitation persists even when enhanced with signature representation.
In the absence of a more meaningful vector representation, we diminish this problem by changing the Ln.~\ref{alg:dtw_end_for} of Alg.~\ref{alg:vec_inferencePS_DTW} to $\mathbb{P}_\goals[g] \gets \mathbb{P}_\goals[g] + \frac{p-\mathbb{P}_\goals[g]}{n_{g}+1}, n_g \gets n_g+1$, where $n_g$ is a counter to keep in track the incremental mean computation.
The mean of the Top-k signatures favors the hypothesis with the greatest number of similarities to the observations.

We follow the same empirical grid search presented in Sec.~\ref{sec:experiments} on $5$ of $12$ environments to find optimal merge and prune thresholds.
In it, we observe that the best values for merge and prune remain zero (see supplementary material for full ablation test results). 
Unlike continuous environments, results for the discrete environments deteriorated as values exceeded $0$.
We hypothesize that this behavior comes from the fact that path signatures use geometrical information to semantically encode different trajectories.
Yet, discrete environments states are either:
(i) encode information in a non-geometrical manner (e.g., one-hot encoding where values represent categorical states without any inherent spatial or geometric relationships), or
(ii) represent transitions that are abrupt and lack continuity, making it difficult for path signatures to capture meaningful patterns.
Consequently, path signatures in discrete spaces produce highly dissimilar representations for each state transition, reducing the effectiveness of pruning operations.
Furthermore, discrete state spaces are finite, which increases the probability of shared trajectory prefixes, limiting the constraint on possible paths and, thus, reducing merging operations.

Nevertheless, the performance reported in Tab.~\ref{tab:planner_compare}b is competitive with other goal recognition approaches in discrete settings, underscoring the robustness and versatility of \abbrev{}.
Fig.~\ref{fig:comp_over_observations}b compares the approaches by their PPV (and its margin of error with a confidence level of 95$\%$), at $30\%$, $50\%$, $70\%$, and $100\%$ of their respective full observation.
Unlike the continuous relative, \abbrev{} and \abbrev+DTW slightly outperform other methods across all stages of observability.
However, the differences between both are not statistically significant, given the previously mentioned vectorization issues.
\section{Related Work}
\label{sec:related_work}

Recent approaches to online goal recognition infer goals by measuring the error between a reference trajectory and the observed state using Euclidean distance.
\citeauthor{fitzpatrick2021behaviour} extend the previous state-of-the-art~\cite{kaminka2018plan}, which relies on consecutive planner calls for each new observation in a scenario while ignoring obstacles. 
Such approaches are unsuitable for real-time applications since relying on calls to a planner incurs an unacceptable computational cost~\cite{jain2021optimal}. 
\citeauthor{tesch2023online} show that a representative number of suboptimal trajectories is sufficient to perform goal inference without relying on consecutive planner call at each observation.
\citet{AmadoAiresPereiraEtAl2019a,AmadoMeneguzzi2020,AmadoPereiraMeneguzzi2023,fang2023real,AmadoMirskyMeneguzzi2022,WenAmado2025,SerinaChiariGereviniEtAl2025,Elhadad2026} leverage machine learning to avoid using planner calls, allowing real-time inference. 
Unlike our approach, they require pre-existing datasets to train the model.
\section{Conclusion}
\label{sec:conclusion}

In this paper, we introduced \abbrev, a novel data-driven framework for recognizing goals from observation sequences.
Unlike traditional approaches, \abbrev does not rely on optimality assumptions or costly planner interactions.
It leverages path signatures, a compact and expressive representation of trajectories, to compare observed behavior against sampled trajectories in a principled and efficient way.

Extensive evaluation on standard benchmarks shows that \abbrev outperforms existing state-of-the-art methods in both precision and planning efficiency, while maintaining strong generalization across scenarios and parameter settings.
While better sampling strategies can be easily incorporated to improve recognition accuracy, key future work consists of developing more efficient ways to find the merge and prune hyperparameters.
Similarly, improvement in discrete domains requires better vector encoding techniques. 

\abbrev assumes access to accurate projection functions for trajectory comparison.
In real-world settings, where noise or high dimensional observations (e.g., raw sensor data or images) are more prevalent, dynamic time warping becomes necessary.
\abbrev{}+DTW does achieve state-of-the-art accuracy, but DTW's added computational cost limits its applicability.
In future work, we plan to explore approximate or differentiable variants of DTW that preserve alignment quality while reducing runtime, making it more accessible for real-time goal recognition.

\section*{Contribution Statement}
Douglas Tesch and Nathan Gavenski have contributed equally to this work.

\section*{Acknowledgments}
This work was partially supported by UK Research and Innovation [grant number EP/S023356/1], in the UKRI Centre for Doctoral Training in Safe and Trusted Artificial Intelligence  (\url{www.safeandtrustedai.org}).

This study was financed in part by the Coordena\c{c}\~ao de Aperfei\c{c}oamento de Pessoal de N\'ivel Superior Brasil (CAPES) - Finance Code 001.

\bibliographystyle{named}
\bibliography{ref}

\clearpage
\appendix
\section{Complexity of \abbrev+DTW}
Regarding complexity, the lookup process in $\tree$ is $\mathcal{O}(1)$ and computing $\signatures{\observations}$ is $\mathcal{O}(\vert\observations\vert \cdot d^k)$, where $\vert\observations\vert$ is the number of time steps in $\observations$, $d$ is the number of dimensions in the agent's observations, and $k$ is the depth of the path signature required by \abbrev.
Therefore, the online inference without $\Call{DTW}{}$ is $\mathcal{O}(|\tree| \cdot \vert\observations\vert \cdot d^k)$, where $\treewidth{\tree}$ is the width of $\tree$.
Thus, it is vital to keep $\tree$ thin by merging its nodes, and finding the right trade-off between expressiveness of higher values of $k$ and efficiency with lower values of $k$ since the sizes $t$ and $d$ are constants.
Yet, $\Call{DTW}{}$'s complexity is $\mathcal{O}(\vert\observations\vert \cdot $\treeheight{\tree}$)$, where $\treeheight{\tree}$ is the size of the longest branch in $\tree$.
By using $\Call{DTW}{}$ in the inference method, \abbrev's complexity increases to $\mathcal{O}\left(\vert\observations\vert \cdot \left( d^k + \treewidth{\tree} \cdot \treeheight{\tree}\right)\right)$, making it vital to also reduce the number of nodes per branch.

\section{Path Signatures}

In the main work, we briefly defined a trajectory $\trajectory$, in which each state is a vector $\vv{v}$ in $\mathbb{R}^d$, and how to compute the path signature $\beta(\trajectory)_{1, n}^{i_1, \cdots. i_k}$, where $n$ is the length of the trajectory, and $i_1, \cdots, i_k \in \{1, \cdots, d \}$ ($k>1$) are indices to elements in $\vv{v}$.
We now briefly explain path signatures and provide some additional information on the process of computing a path signature, its trajectory tree $TT$, and the intuition behind it.
For a more in-depth examination on this topic, we refer the reader to~\citeauthor{chevyrey2016signature}~\shortcite{chevyrey2016signature}.

\subsection{Computing the Path Signature} \label{sec:computing_path_signatures}

Given a trajectory $\trajectory$ and a function $f$ that interpolates $\trajectory$ into a continuous map $f: \mathbb{R} \rightarrow \mathbb{R}$, the integral of the trajectory against $f$ can be defined as:
\normalsize
\begin{equation}
    \int_1^n  f(\trajectory_t)d\trajectory_t = \int_1^n f(\trajectory_t)\dot{\trajectory}_tdt,
\end{equation}
\normalsize
where $\dot(\trajectory)_t = \frac{d\trajectory}{dt}$ for any time $t \in [1, n]$.
It is important to note that $f(\trajectory_t)$ is a real-valued path defined on $[1, n]$, which can be considered the integral of a trajectory $\trajectory$.
Moreover, if we consider that $f(\trajectory_t) = 1$ for all $t \in [1, n]$, then the path integral of $f$ against any trajectory $\trajectory: [1, n] \rightarrow \mathbb{R}$ is simply the increment of $\trajectory$:
\normalsize
\begin{equation}
    \int_1^n d\trajectory_t = \int_1^n \dot{\trajectory}dt = \trajectory_n - \trajectory_1.    
\end{equation}
\normalsize

Therefore, by assuming that $\beta$ is a function of real-valued paths, we can define the signature for any single index $i_k \in \{1, \cdots, d\}$ as:
\normalsize
\begin{equation}  \label{eq:single_index}
    \beta(\trajectory)^{i_k}_{1, n} = \int_{1 < s \leqslant n} d\trajectory^{i_k}_s = \trajectory^{i_k}_n - \trajectory^{i_k}_1, 
\end{equation}
\normalsize
which is the increment of the $i_k$-th dimension of the path.
If we move to any pair of indexes $i_k, j_k \in \{1, \cdots, d\}$, we have to consider the double-iterated integral:
\normalsize
\begin{equation} \label{eq:double_index}
    \beta(\trajectory)^{i_k, j_k}_{1, n} = \int_{1 < s \leqslant n} \beta(\trajectory)^{i_k}_{1, s}d\trajectory^{j_k}_s = \int_{1 < r \leqslant s \leqslant n} d\trajectory^{i_k}_t d\trajectory^{j_k}_s,
\end{equation}
\normalsize
where $\beta(\trajectory)^{i_k}_{1, s}$ is given by Eq~\ref{eq:single_index}.
Considering that $\beta(\trajectory)^{i_k, j_k}_{1, n}$ continues to be a real-valued path, then we can define recursively the signature function for any number of indexes $k \geqslant 1$ in the collection of indexes $i_1, \cdots, i_k, \in \left\{ 1, \cdots, d \right\}$ as:
\normalsize
\begin{equation} \label{eq:signature}
    \beta(\trajectory)^{i_1,\cdots,i_k}_{1, n} =
    \int_{1 < s \leqslant n} \beta(\trajectory)^{i_1, \cdots, i_{k-1}}_{1, s} d\trajectory^{i_k}_s,
\end{equation}
\normalsize
which in the main work is Equation $1$.
We want to highlight that $k$ is the depth of the signature up to which the signature is generated (not its length).
At each level $i \leqslant k$, features of length $i$ are generated from the dimensions $D$ according to Eq~\ref{eq:signature} to produce the terms of the signature.
Figure~\ref{fig:dictionary} shows all possible terms for a trajectory $\trajectory: [1, n] \rightarrow \mathbb{R}^d$ for different levels.
For example, a signature with depth $2$ will have all the terms in the levels $i=0,1,2$. In a vector with $d$ features, we can construct one feature of length $0$, $d$ features of length $1$, and $d^2$ features of length $2$, giving $1+d+d^2$ features in total (i.e., the number of terms in the signature). 
The length of a signature with a vector size $d$ and depth $k$ is:
\normalsize
\begin{equation}\label{eq:signature_size}
    \sum_{i=0}^{k}d^i=\frac{d^{k+1}-1}{d-1}.
\end{equation}
\normalsize
We observe that signatures can be computed for any depth $k$, and are not restricted to $k \leqslant d$.

\begin{figure}[h!]
   \centering 
   \begin{tikzpicture}[]
    
    \matrix (m) [
        matrix of nodes,
        nodes in empty cells,
        row sep=-1pt,
        column sep=3pt,
        nodes={minimum width=20pt, minimum height=20pt}
    ]{
        1 & $\beta(\trajectory)^1_{1, n}$ & $\beta(\trajectory)^{1, 1}_{1, n}$ & $\beta(\trajectory)^{1, 1, 1}_{1, n}$ & $\cdots$ & $\beta(\trajectory)^{1, 1, \cdots, 1}_{1, n}$ \\
          & $\beta(\trajectory)^2_{1, n}$ & $\beta(\trajectory)^{1, 2}_{1, n}$ & $\beta(\trajectory)^{1, 1, 2}_{1, n}$ & $\cdots$ & $\beta(\trajectory)^{1, 1, \cdots, 2}_{1, n}$ \\
          & $\vdots$               & $\vdots$                    & $\vdots$                       &          & $\vdots$ \\
          & $\beta(\trajectory)^d_{1, n}$ & $\beta(\trajectory)^{1, d}_{1, n}$ & $\beta(\trajectory)^{1, 1, d}_{1, n}$ & $\cdots$ & $\beta(\trajectory)^{1, 1 \cdots, d}_{1, n}$ \\
          &                        & $\beta(\trajectory)^{2, 1}_{1, n}$ & $\beta(\trajectory)^{1, 2, 1}_{1, n}$ & $\cdots$ & $\beta(\trajectory)^{i_1, i_2, \cdots, i_k}_{1, n}$ \\
          &                        & $\vdots$                    & $\vdots$                       &          & $\vdots$ \\
          &                        & $\beta(\trajectory)^{d, 1}_{1, n}$ & $\beta(\trajectory)^{d, d, 1}_{1, n}$ & $\cdots$ & $\beta(\trajectory)^{d, d, \cdots, 1}_{1, n}$ \\
          &                        & $\beta(\trajectory)^{d, 2}_{1, n}$ & $\beta(\trajectory)^{d, d, 2}_{1, n}$ & $\cdots$ & $\beta(\trajectory)^{d, d, \cdots, 2}_{1, n}$ \\
          &                        & $\vdots$                    & $\vdots$                       &          & $\vdots$ \\
          &                        & $\beta(\trajectory)^{d, d}_{1, n}$ & $\beta(\trajectory)^{d, d, d}_{1, n}$ & $\cdots$ & $\beta(\trajectory)^{d, d, \cdots, d}_{1, n}$ \\
    };
    
    \draw [->, >=latex]
        ([xshift=-8pt, yshift=2pt]m-1-1.north west) -- 
        ([yshift=2pt]m-1-6.north east) {};
    \node [above=2pt] at (m-1-1.north)  {\scriptsize $0$};
    \node [above=2pt] at (m-1-2.north)  {\scriptsize $1$};
    \node [above=2pt] at (m-1-3.north)  {\scriptsize $2$};
    \node [above=2pt] at (m-1-4.north)  {\scriptsize $3$};
    \node [above=2pt] at (m-1-5.north)  {\scriptsize $\cdots$};
    \node [above=2pt] at (m-1-6.north)  {\scriptsize $k$};

    \draw [->, >=latex]
        ([yshift=10pt]m-1-1.north west) --
        (m-10-1.south west) node[midway, left=2pt] {\rotatebox{90}{$d^i\text{ items}$}};
    \node [xshift=-6pt, above=2pt] at (m-1-1.north west) {\scriptsize $i$};
    \node [xshift=10pt, above=12pt] at (m-1-4.north west) {\scriptsize depth};
\end{tikzpicture}
   \caption{collection of signatures for $\trajectory$, where $k \in [0, \infty)$.}
   \label{fig:dictionary}
\end{figure}

\subsection{Path Signatures Numerical Example}

\begin{figure}[ht]
    \centering
        \begin{tikzpicture}
        \begin{axis}[
            width=10cm,
            height=7cm,
            xlabel={$\trajectory^1$},
            ylabel={$\trajectory^2$},
            xmin=5, xmax=19,
            ymin=0, ymax=350,
            axis lines=left,
            legend style={at={(0.5,-0.15)}, anchor=north},
            samples=200,
        ]
            \addplot[blue, thick, dashed, domain=0:100] ({5 + x}, {(5 + x)^2});
            \addplot[blue, thick, domain=1:10] ({5 + x}, {(5 + x)^2});
            
            \addplot[dashed, gray, thick] coordinates {(6, 0) (6, 36)};
            \addplot[only marks, mark=*] coordinates {(6, 36)};
            
            \addplot[dashed, gray, thick] coordinates {(15, 0) (15, 225)};
            \addplot[only marks, mark=*] coordinates {(15, 225)};
        \end{axis}
    \end{tikzpicture}
    \caption{Trajectory $\trajectory: [1, 10] \rightarrow \mathbb{R}^2$.}
    \label{fig:example}
\end{figure}
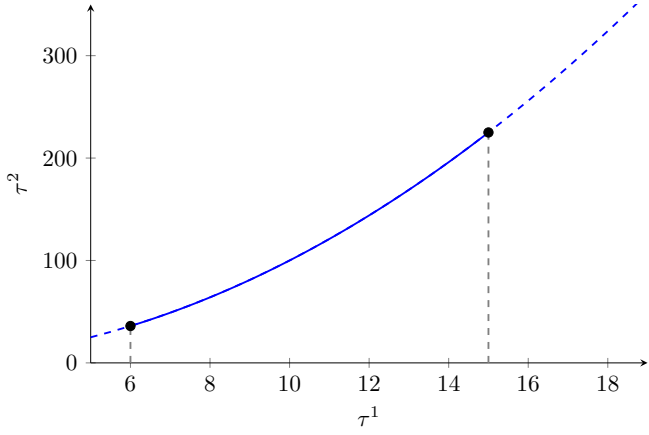

We now consider a trajectory $\trajectory$ with two two-dimensional states $\{\trajectory_t^1, \trajectory_t^2\}$, and the set of multi-indexes $W = \left\{\right. \left( i_1, \cdots, i_k\right) \mid k \geqslant 1, i_1, \cdots, i_k \in \left\{1, 2\right\} \left.\right\}$, which is the set of all finite sequences of $1$'s and $2$'s.
Given the trajectory $\trajectory: [1, 10] \rightarrow \mathbb{R}^2$ illustrated in Figure~\ref{fig:example}, where the path function for $\trajectory$ follows:

\begin{equation}
    \begin{aligned}
        \trajectory_t &= \{ \trajectory_t^1, \trajectory_t^2 \} = \{ 5 + t, (5 + t)^2 \mid t \in [1, 10] \} \\
    \end{aligned}
\end{equation}

For the depth $k$ desired, the computation of the signature is as shown in Figure~\ref{fig:step-by-step}.
For example, given states in $\trajectory$ are two-dimensional ($d=2$), the path signature for $\trajectory$ width depth $k=2$ will have the $\frac{d^{k+1}-1}{d-1}=\frac{2^{3}-1}{1}=7$ terms in the vector $\beta(\trajectory)_{1, 10} = \left[\right. 1,  9, 189, 40.5, 970.5, 730.5, 17860.5 \left.\right]$.

\begin{figure}[htb!]
    \centering
    \begin{tikzpicture}[scale=1.5, node distance=2mm, every node/.style={align=center, anchor=west}]
    \node[align=center] (k3) {
        $\begin{aligned}
            \beta(\trajectory)^{1, 1, 1}_{1, 10} &= \int\!\!\!\!\int\!\!\!\!\int_{1 < t_1 \leqslant t_2 \leqslant t_3 \leqslant 10} d\trajectory^1_{t_1}d\trajectory^1_{t_2}d\trajectory^1_{t_3} = \int_1^{t_1} \left[ \int_1^{t_2} \left[ \int_1^{t_3} dt_1 \right] dt_2 \right] dt_3 = 121.5 \\
            &\vdots
        \end{aligned}$
    };

    \node[align=center, above=of k3] (k2) {
        $\begin{aligned}
            \beta(\trajectory)^{1, 1}_{1, 10} &= \int\!\!\!\!\int_{1 < t_1 \leqslant t_2 \leqslant 10} d\trajectory^1_{t_1}, d\trajectory^1_{t_2} = \int_1^{10} \left[ \int_1^{t_2} dt_1 \right] dt_2 = 40.5 \\
            \beta(\trajectory)^{1, 2}_{1, 10} &= \int\!\!\!\!\int_{1 < t_1 \leqslant t_2 \leqslant 10} d\trajectory^1_{t_1}, d\trajectory^2_{t_2} = \int_1^{10} \left[ \int_1^{t_2} dt_1 \right] 2(5 + t_2)dt_2 = 970.5 \\
            \beta(\trajectory)^{2, 1}_{1, 10} &= \int\!\!\!\!\int_{1 < t_1 \leqslant t_2 \leqslant 10} d\trajectory^2_{t_1}, d\trajectory^1_{t_2} = \int_1^{10} \left[ \int_1^{t_2} 2(5+t)dt_1 \right] dt_2 = 730.5 \\
            \beta(\trajectory)^{2, 2}_{1, 10} &= \int\!\!\!\!\int_{1 < t_1 \leqslant t_2 \leqslant 10} d\trajectory^2_{t_1}, d\trajectory^2_{t_2} = \int_1^{10} \left[ \int_1^{t_2} 2(5+t)dt_1 \right] 2(5+t)dt_1 = 17,860.5
        \end{aligned}$
    };
    
    \node[above=of k2.north west, anchor=south west] (k1) {
        $\begin{aligned}
            \beta(\trajectory)^1_{1, 10} &= \int_{1 < t \leqslant 10} dt = \trajectory^1_{10} - \trajectory^1_1 = 9\\
            \beta(\trajectory)^1_{1, 10} &= \int_{1 < t \leqslant 10} dt = \trajectory^2_{10} - \trajectory^2_1 = 189
        \end{aligned}$
    };
    
    \node[above=of k1.north west, anchor=south west] (k0) {$\beta(\trajectory)^{\o}_{1, n} = 1$};

    \draw[-, >=latex] ([yshift=-2pt]$(current bounding box.west |- k0.south west)$) -- ([yshift=-2pt]$(current bounding box.east |- k0.south east)$);
    \draw[-, >=latex] ([yshift=-2pt]$(current bounding box.west |- k1.south west)$) -- ([yshift=-2pt]$(current bounding box.east |- k1.south east)$);
    \draw[-, >=latex] ([yshift=-2pt]$(current bounding box.west |- k2.south west)$) -- ([yshift=-2pt]$(current bounding box.east |- k2.south east)$);

    \node[anchor=east] at ([xshift=-2pt]k0.west) {\rotatebox{90}{\scriptsize $k = 0$}};
    \node[anchor=east] at ([xshift=-2pt]k1.west) {\rotatebox{90}{\scriptsize $k = 1$ -- Eq. 3}};
    \node[anchor=east] at ([xshift=-2pt]k2.west) {\rotatebox{90}{\scriptsize $k = 2$ -- Eq. 4}};
    \node[anchor=east] at ([xshift=-2pt]k3.west) {\rotatebox{90}{\scriptsize $k \geqslant 3$ -- Eq. 5}};

\end{tikzpicture}
    \caption{Step-by-step computation of a path signature}
    \label{fig:step-by-step}
\end{figure}

\subsection{Path Signature Trees}

Path signatures have a uniqueness property: no two trajectories $\trajectory$ and $\trajectory'$ of bounded variation have the same signature unless the trajectories are identical.
Leveraging this property, we can represent a trajectory as a tree $\tree$ constructed from the path signatures of its partial trajectories.
Consider a trajectory $\trajectory$; by computing the path signature for each partial trajectory $\trajectory_{1, t}$ (i.e., the trajectory from the first state up to time $t$), we obtain a sequence of path signatures that form a linear tree.
Each node in this branch corresponds to a path signature that encodes the trajectory of increasing length.
We remind the reader that path signatures are of fixed lengths due to their invariance to the trajectory size.

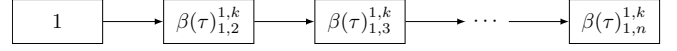
\begin{figure}[h!tb]
    \centering
        \begin{tikzpicture}[node distance=1cm]
        \node[tree_node] (zeroth) at (0, 0) {$1$};
        \node[tree_node] (first) [right=of zeroth] {$\beta(\trajectory)^{1,k}_{1,2}$};
        \node[tree_node] (second) [right=of first] {$\beta(\trajectory)^{1,k}_{1,3}$};
        \node (third) [right=of second] {$\cdots$};
        \node[tree_node] (fourth) [right=of third] {$\beta(\trajectory)^{1,k}_{1,n}$};

        \draw [arrow] (zeroth.east) -- (first.west);
        \draw [arrow] (first.east) -- (second.west);
        \draw [arrow] (second.east) -- (third.west);
        \draw [arrow] (third.east) -- (fourth.west);
    \end{tikzpicture}
    \caption{Trajectory tree for a single trajectory.}
    \label{fig:tt_single}
\end{figure}

For a trajectory of length $n$, we compute $n-1$ path signatures up to depth $k$, one for each partial trajectory.
This sequence is represented as:
\begin{equation}
    \left[1, \beta(\trajectory)_{1, 2}^{1, k}, \beta(\trajectory)_{1, 3}^{1, k}, \cdots, \beta(\trajectory)_{1, n}^{1, k} \right],
\end{equation}
where the first entry (the zeroth term) is always $1$ since no path signatures exist for a trajectory of length 1, and the final term corresponds to the path signature of the whole trajectory (displayed in Figure~\ref{fig:step-by-step}).
For any single trajectory, the path signature tree would be a single branch with the root node being the zeroth term and the leaf being the complete trajectory $\trajectory_{1,n}$ path signature (Figure~\ref{fig:tt_single}).

Now consider any set of trajectories $\alltrajectories$ with $| \alltrajectories | > 1$.
Due to the uniqueness property of path signatures, in this case, two trajectories share a node if and only if their partial trajectories are identical up to that point -- that is, $\trajectory_{1, t} = \trajectory'_{1, t}$ for all $t$ such that $\trajectory_t = \trajectory'_t$).
Thus, given a set of trajectories $\alltrajectories$, it is possible to build a trajectory tree, where the maximum \textit{height} is the size of the longest trajectory $\max_{\trajectory \in \alltrajectories} |\trajectory|$, and the maximum \textit{width} is the number of trajectories $|\alltrajectories|$ if each trajectory is unique.

\subsection{Trajectory Tree Numerical Example}

\begin{figure}[b]
    \centering
    \begin{tikzpicture}
\begin{axis}[
    width=12cm,
    height=8cm,
    xlabel={$\trajectory^1$},
    ylabel={$\trajectory^2$},
    xmin=0.5, xmax=4.5,
    ymin=-0.5, ymax=4.5,
    legend style={at={(1.02,1)}, anchor=north west},
    axis lines=left,
    samples=200,
]

\addplot[cb_blue, thick, domain=1:4] {2};
\addlegendentry{$\trajectory$}
\addplot[cb_blue, thick, dashed, domain=0:1, forget plot] {2};
\addplot[cb_blue, thick, dashed, domain=4:5, forget plot] {2};

\addplot[cb_orange, thick, domain=1:4] {max(x, 2)};
\addlegendentry{$\trajectory'$}
\addplot[cb_orange, thick, dashed, domain=0:1, forget plot] {max(x, 2)};
\addplot[cb_orange, thick, dashed, domain=4:5, forget plot] {max(x, 2)};

\addplot[cb_green, thick, domain=1:4] {min(2, x-1)};
\addlegendentry{$\trajectory''$}
\addplot[cb_green, thick, dashed, domain=0:1] {min(2, x-1)};
\addplot[cb_green, thick, dashed, domain=4:5] {min(2, x-1)};

\end{axis}
\end{tikzpicture}
    \caption{Trajectories for $\trajectory, \trajectory', \text{ and } \trajectory''$.}
    \label{fig:tree_example}
\end{figure}

We now illustrate this with three example trajectories $\trajectory$, $\trajectory'$, and $\trajectory''$, all defined over the same time interval with two-dimensional states and path signatures depth $k=2$.
The trajectories, shown in Figure~\ref{fig:tree_example}, are defined as follows:
\begin{equation}
    \begin{split}
        &\trajectory = \{ \trajectory^1_t, \trajectory^2_t \} = \{ t, 2 \mid t \in [1, 4] \}\\    
        &\trajectory' =  \{ \trajectory'^1_t, \trajectory'^2_t \} = \{ t, \max(t, 2) \mid t \in [1, 4] \} \\
        &\trajectory'' =  \{ \trajectory''^1_t, \trajectory''^2_t \} = \{ t, \min(2, t-1) \mid t \in [1, 4] \}
    \end{split}    
\end{equation}

Building the trajectory tree for these three trajectories yields the structure shown in Figure~\ref{fig:enter-label}.

\begin{figure}[h!tbp]
    \centering
        \begin{tikzpicture}[node distance=1cm]
        \node[tree_node] (zeroth) at (0, 0) {$1$};

        \node[tree_node] (first) [below=of zeroth] {$[1.0, 0.0, 0.5, 0.0, 0.0, 0.0, ]$};
        \node[tree_node] (second) [below=of first] {$[2.0, 0.0, 2., 0.0, 0.0, 0.0]$};
        \node[tree_node] (third) [below=of second] {$[3.0, 0.0, 4.5, 0.0, 0.0, 0.0]$};

        \node[tree_node] (second_2) [left=of second] {$[2.0, 1.0, 2.0, 1.5, 0.5., 0.5]$};
        \node[tree_node] (third_2) [below=of second_2] {$[3.0, 2.0, 4.5, 4.0, 2.0, 2.0]$};

        \node[tree_node] (first_3) [right=of first] {$[1.0, 1.0, 0.5, 0.5, 0.5, 0.5]$};
        \node[tree_node] (second_3) [below=of first_3] {$[2.0, 2.0, 2.0, 2.0, 2.0., 2.0]$};
        \node[tree_node] (third_3) [below=of second_3] {$[3.0, 2.0, 4.5, 2.0, 4.0, 2.0]$};

        \draw [arrow, cb_blue] (zeroth.south) -- (first.north);
        \draw [arrow, cb_blue] (first.south) -- (second.north);
        \draw [arrow, cb_blue] (second.south) -- (third.north);

        \draw [arrow, cb_orange] ([yshift=-0.5cm]first.south) -| (second_2.north);
        \draw [arrow, cb_orange] (second_2.south) -- (third_2.north);
        
        \draw [arrow, cb_green] ([yshift=-0.5cm]zeroth.south) -| (first_3.north);
        \draw [arrow, cb_green] (first_3.south) -- (second_3.north);
        \draw [arrow, cb_green] (second_3.south) -- (third_3.north);
        
        \threecolorcircle{cb_blue}{cb_orange}{cb_green}{zeroth};
        \twocolorcircle{cb_blue}{cb_orange}{first};
        \onecolorcircle{cb_blue}{second};
        \onecolorcircle{cb_blue}{third};
        \onecolorcircle{cb_orange}{second_2};
        \onecolorcircle{cb_orange}{third_2};
        \onecolorcircle{cb_green}{first_3};
        \onecolorcircle{cb_green}{second_3};
        \onecolorcircle{cb_green}{third_3};
    \end{tikzpicture}
    \caption{Trajectories tree for $\trajectory, \trajectory', \text{ and } \trajectory''$.}
    \label{fig:enter-label}
\end{figure}

\noindent
In this example, the colored dots represent each trajectory, e.g., the zeroth node has all three trajectories (all three colored dots).
It is essential to observe that given the uniqueness property from path signatures, even if $\trajectory$ and $\trajectory''$ finish in the exact location (in our main work, we refer to this as the same goal $g \in \goals$), they diverge immediately after the root node since their trajectories are distinct.
This illustrates how the path signature tree distinguishes between trajectories even if their endpoints coincide.
In contrast, $\trajectory$ and $\trajectory'$ share identical states for the first two time steps and therefore follow a common branch beyond the root.
This shared structure is captured by the path signature tree, reflecting the underlying similarity in the early portions of their trajectories.

\subsection{Pruning and Merging Motivation}

\begin{figure}[b]
    \centering
    \begin{tikzpicture}
\begin{axis}[
    width=12cm,
    height=8cm,
    xlabel={$\trajectory^1$},
    ylabel={$\trajectory^2$},
    xmin=0.5, xmax=4.5,
    ymin=-0.5, ymax=4.5,
    legend style={at={(0.02,1)}, anchor=north west},
    axis lines=left,
    samples=200,
]

\addplot[cb_blue, thick, domain=1:4] {2};
\addlegendentry{$\trajectory$}
\addplot[cb_blue, thick, dashed, domain=0:1, forget plot] {2};
\addplot[cb_blue, thick, dashed, domain=4:5, forget plot] {2};

\addplot[cb_orange, thick, domain=1:4] {max(x, 2)};
\addlegendentry{$\trajectory'$}
\addplot[cb_orange, thick, dashed, domain=0:1, forget plot] {max(x, 2)};
\addplot[cb_orange, thick, dashed, domain=4:5, forget plot] {max(x, 2)};

\addplot[cb_green, thick, domain=1:4] {min(2, x-1)};
\addlegendentry{$\trajectory''$}
\addplot[cb_green, thick, dashed, domain=0:1, forget plot] {min(2, x-1)};
\addplot[cb_green, thick, dashed, domain=4:5, forget plot] {min(2, x-1)};

\addplot[cb_purple, thick, domain=1:4] {0.1 * exp(-((x - 2)^2) / (2 * 0.158^2)) + 2};
\addlegendentry{$\trajectory'''$}
\addplot[cb_purple, thick, dashed, domain=0:1] {0.1 * exp(-((x - 2)^2) / (2 * 0.158^2)) + 2};
\addplot[cb_purple, thick, dashed, domain=4:5] {0.1 * exp(-((x - 2)^2) / (2 * 0.158^2)) + 2};
\end{axis}
\end{tikzpicture}
    \caption{Trajectories for $\trajectory, \trajectory', \trajectory'', \text{ and } \trajectory'''$.}
    \label{fig:new_trajectory}
\end{figure}

We now introduce a fourth trajectory $\trajectory'''$, following the same setting from the previous examples, displayed in Figure~\ref{fig:new_trajectory} and defined as:
\begin{equation}
    \trajectory''' = \{ \trajectory'''^1_t, \trajectory'''^2_t \} = \{ t, 0.1 \cdot \exp \left( -\frac{(t-2)^2}{2 \!\cdot\! 0.158^2} \right) + 2 \}
\end{equation}

Adding this new trajectory to the trajectory tree, despite its only slight variation from $\trajectory$ at $t = 2$, results in the updated structure shown in Figure~\ref{fig:new_tree}.
\begin{figure}[h!btp]
    \centering
    \begin{tikzpicture}[node distance=1cm]

\node[tree_node] (zeroth) at (0, 0) {$1$};

\node[tree_node] (first)  [below=of zeroth] {$[1.0, 0.0, 0.5, 0.0, 0.0, 0.0]$};
\node[tree_node] (second) [below=of first]  {$[2.0, 0.0, 2.0, 0.0, 0.0, 0.0]$};
\node[tree_node] (third)  [below=of second] {$[3.0, 0.0, 4.5, 0.0, 0.0, 0.0]$};

\node[tree_node] (second_2) [left=of second] {$[2.0, 1.0, 2.0, 1.5, 0.5, 0.5]$};
\node[tree_node] (third_2)  [below=of second_2] {$[3.0, 2.0, 4.5, 4.0, 2.0, 2.0]$};

\node[tree_node] (first_3)  [right=of first] {$[1.0, 1.0, 0.5, 0.5, 0.5, 0.5]$};
\node[tree_node] (second_3) [below=of first_3] {$[2.0, 2.0, 2.0, 2.0, 2.0, 2.0]$};
\node[tree_node] (third_3)  [below=of second_3] {$[3.0, 2.0, 4.5, 2.0, 4.0, 2.0]$};

\node[tree_node] (first_4)  [right=of first_3] {$[1.000, 0.100, 0.500, 0.050, 0.050, 0.005]$};
\node[tree_node] (second_4) [below=of first_4] {$[2.000, 0.000, 2.000, -0.100, 0.100, 0.000]$};
\node[tree_node] (third_4)  [below=of second_4] {$[3.000, 0.000, 4.500, -0.100, 0.100, 0.000]$};

\draw [arrow, cb_blue]   (zeroth.south) -- (first.north);
\draw [arrow, cb_blue]   (first.south)  -- (second.north);
\draw [arrow, cb_blue]   (second.south) -- (third.north);

\draw [arrow, cb_orange] ([yshift=-0.5cm]first.south) -| (second_2.north);
\draw [arrow, cb_orange] (second_2.south) -- (third_2.north);

\draw [arrow, cb_green]  ([yshift=-0.6cm]zeroth.south) -| (first_3.north);
\draw [arrow, cb_green]  (first_3.south) -- (second_3.north);
\draw [arrow, cb_green]  (second_3.south) -- (third_3.north);

\draw [arrow, cb_purple] ([yshift=-0.45cm]zeroth.south) -| (first_4.north);
\draw [arrow, cb_purple] (first_4.south) -- (second_4.north);
\draw [arrow, cb_purple] (second_4.south) -- (third_4.north);

\fourcolorcircletree{cb_blue}{cb_orange}{cb_green}{cb_purple}{zeroth};
\twocolorcircletree{cb_blue}{cb_orange}{first};
\onecolorcircletree{cb_blue}{second};
\onecolorcircletree{cb_blue}{third};
\onecolorcircletree{cb_orange}{second_2};
\onecolorcircletree{cb_orange}{third_2};
\onecolorcircletree{cb_green}{first_3};
\onecolorcircletree{cb_green}{second_3};
\onecolorcircletree{cb_green}{third_3};
\onecolorcircletree{cb_purple}{first_4};
\onecolorcircletree{cb_purple}{second_4};
\onecolorcircletree{cb_purple}{third_4};

\end{tikzpicture}
    \caption{Trajectory tree for $\trajectory, \trajectory', \trajectory'', \text{ and } \trajectory'''$.}
    \label{fig:new_tree}
\end{figure}

Although $\trajectory'''$ closely resembles $\trajectory$ for most of its duration, it exhibits a slight deviation at $t = 2$.
Due to the uniqueness property of path signatures, this slight variation results in an entirely new branch in the tree -- highlighting how path signature trees encode fine-grained differences between trajectories.
Apart from the shared root node, $\trajectory'''$ does not overlap with any of the previously defined trajectories.
This example underscores how even subtle variation, such as the deviation of $\trajectory'''$ from $\trajectory$ at $t=2$, can produce entirely new branches in the trajectory tree due to the sensitivity of path signatures.
While this property is valuable for distinguishing trajectory structures, it also results in unnecessarily complex trees with redundant or overly fine-trained branches.
In the context of goal recognition, such complexity can hinder both interpretability and computation efficiency.
To address this, our main work introduces pruning and merging mechanisms that simplify the trajectory tree by removing insignificant structural differences while retaining the distinctions critical for goal inference.
Notably, after applying these operations, the resulting tree often contains stable intermediate nodes that consistently appear across different trajectories.
These nodes serve as functional analogues to \textit{landmarks}, informative subgoals or waypoints that offer strong cues about the agent's eventual objective.
This connection allows us to reason about goals not only based on endpoints but also by leveraging these shared structural checkpoints.

\section{Experimental Details}

We conducted the experiments using a $2.2$GHz $24$-core CPU with $24$GB RAM, running Ubuntu $20.04$. 
We use the `signatory' package to compute each path signature~\cite{kidger2021signatory}, and the sample function from \citet{tesch2023online}.
The path signatures are all with depth $2$, i.e., $k = 2$.
The code for all experiments is available at: \url{https://github.com/douglasat/GRPS}.

\section{Dataset}

The dataset simulates a robot with a two-wheeled motor and one unidirectional wheel traveling between an initial and a goal point. 
Each map in the dataset contains eight randomly sampled points, $p_1$ to $p_8$, that can be used as initial and goal points.
Each point contains $x$ and $y$ coordinates, as well as an orientation in radians. 
For each map, the dataset contains $56$ problems, which comprise all pairs of initial and goal states.
For example, each map will have one problem where the ground truth is a trajectory from $p_1$ to $p_2$ (whose $x,y$ coordinates we call $g_2$), with $p_2$ to $p_8$ being goal hypotheses, another one with $p_8$ to $p_7$ (\textit{q.v.} $g_7$) with $p_1$ to $p_7$ as goal hypotheses, and so on. 
We use the scenario map Aftershock as a working example in Figure~\ref{fig:map}. 
White represents traversable space, and marked colored points represent potential initial and goal position points. 
Here, we see the randomly selected points in the map, as well as potential trajectories between points $p_n$ from Figure~\ref{fig:map}.
The dataset provides the observations for each recognition problem, assuming the robot travels optimally, i.e., the robot moves while minimizing travel time as a cost function.
Figure~\ref{fig:obser} shows an example of optimal observations ({\color{orange}orange}) and their approximated trajectory ({\color{green} green}).
Although both trajectories look quite similar, there are points where the approximated trajectories are not identical and may display more drastic differences.
We used the dataset available at: \url{https://github.com/douglasat/Vectorial-Goal-Recognition}

\begin{figure}
    \centering
    \begin{subfigure}[t]{0.49\columnwidth}
        \centering
        \includegraphics[width=\columnwidth]{Figures/map.pdf}
        \caption{Aftershock map example.}
        \label{fig:map}
    \end{subfigure}
    \hfill
    \begin{subfigure}[t]{0.49\columnwidth}
        \centering
        \includegraphics[width=\columnwidth]{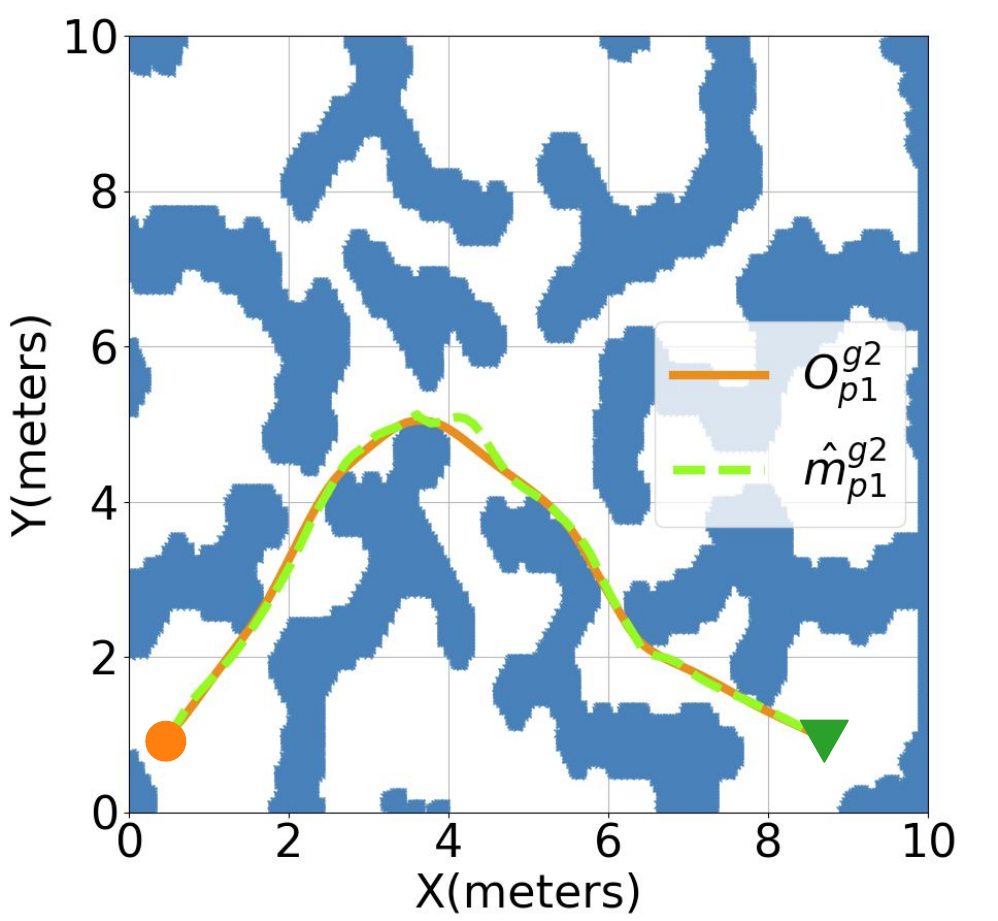}
        \caption{Approximated trajectory $\hat{\trajectory}_{p_1}^{g_2}$ and the observation $O_{p_1}^{g_2}$.}
        \label{fig:obser}
    \end{subfigure}
    \caption{Working dataset examples.}
    \label{fig:trajectory_tree_exemple}
\end{figure}

\section{Partial Observability Algorithm}

We provide a more detailed version of Algorithm 1 from the main work here.
In it, we detail \abbrev{}+DTW procedures for the partial observability setting.

Algorithm assumes that some observations might not reach \abbrev{}+DTW during the online process for some reason, e.g., faulty systems.
Therefore, \abbrev{}+DTW first checks if the new observation $o_t$ is a direct sequence from the last observation $o_{last_t}$ (Line~\ref{alg:partial_check}).
If the sequence of observations is indeed incomplete ($last_t \neq t - 1$), then \abbrev{}+DTW uses an interpolation function $\Call{Interpolate}{}$ to approximate the sequence of observations.
For example, if the current sequence of observation is $\left[\initialstate, o_1, o_3 \right]$ for $t=3$, than $\Call{Interpolate}{}$ returns $\left[\initialstate, o_1, \hat{o}_2, o_3 \right]$, where $\hat{o}_2$ is the approximated obsevation from the interpolation function.
For these experiments, we use Scipy's interpolation function.\footnote{\url{https://docs.scipy.org/doc/scipy/reference/generated/scipy.interpolate.interp1d.html}}
With the new complete sequence of observations up to $t$, \abbrev{}{}+DTW has to retrieve all partial signatures from $last_t$ to $t-1$ (Lines~\ref{alg:partial_signatures_for}-\ref{alg:partial_signatures_for_end}).
That is, for the example $\left[\initialstate, o_1, \hat{o}_2, o_3 \right]$, \abbrev{}+DTW retrieves the path signature $\pathsignature{\observations}^k_{0:2}$, so the set $\signatures{\observations}$ would be $\left\{1, \pathsignature{\observations}^k_{0:1}, \pathsignature{\observations}^k_{0:2}\right\}$.
Afterwards \abbrev{}+DTW stores the last known timestep at $last_t$ and continues the algorithm.

\begin{algorithm}[h!tbp] 
    \scriptsize
    \caption{Partial Observability \abbrev{}+DTW}\label{alg:detailed_dtw}
    \begin{algorithmic}[1]
    \Require $\domain, \goals, \initialstate$ from goal recognition problem
    \Require $\topksize$ number of top trajectories
    \Require $\mergethreshold, \prunethreshold, k$ to build trajectory tree 

    \smallskip
    \Statex \textbf{\method } \Comment{Offline}
    \State $\topktrajectoriesnog \gets [\,]$
    \For{$\textbf{all} \, {\goalstate} \in \goals$} \label{DTW:loop1start}
        \State $\topktrajectories \gets \Call{SampleKTraj}{\domain, \initialstate, \goal, \topksize}$ \label{DTW:function_traj}
        \State $\topktrajectoriesnog \gets \topktrajectoriesnog \cup \topktrajectories$
    \EndFor \label{DTW:loop1stop}
    \State $\tree \gets \Call{trajectoryTree}{\topktrajectoriesnog, \mergethreshold, \prunethreshold, k}$ \Comment{Algorithm 2} \label{DTW:merge_prune}

    \smallskip
    \Statex \textbf{Dynamic Time Warping} \Comment{Online}
    \State $\observations \gets [\,]$
    \State $\signatures{\observations} \gets [\,]$
    \State $last_t \gets 0$
    \When{receives $o_\discretesteptime$ and $o_\discretesteptime \notin \goals$} \label{DTW:online_while}
            \State $\observations \gets \observations \cup o_t$
            \If{$t - 1 \neq last_t$} \label{alg:partial_check}
                \State $\observations \gets \Call{Interpolate}{\observations, t, last_t}$ \label{alg:interpolate}
                \For{$z \in t - last_t - 1$} \label{alg:partial_signatures_for}
                    \State $z \gets z + last_t$
                    \State $\signatures{\observations} \gets \signatures{\observations} \cup \pathsignature{\observations}^k_{0:z}$ 
                \EndFor \label{alg:partial_signatures_for_end}
            \EndIf
            \State $last_t \gets t$
            \State $\signatures{\observations} \gets \signatures{\observations} \cup \pathsignature{\observations}^k_{1:t}$ \label{DTW:O_sig} \Comment{Compute observations signature}
            \State $\mathbb{P}_\goals \gets \{ g \mapsto 0 \mid g \in \goals \}$ \Comment{Initialize probabilities}
            \For{$branch \in \tree$} \label{alg:dtw_for}
                \State $\alignindex \gets \dtw{\signatures{\observations}}{branch}$\label{DTW:index}
                \State $g \gets \text{leaf node of } branch$ \label{DTW:goal}
                \State $p \gets 1 - \exp(\nicefrac{-1}{\frac{1}{|O|}\sum_{i, j \in \delta} \left\| {\beta_O}_i - branch_j \right\|^2)}$\label{DTW:line_compute}
                \State $\mathbb{P}_\goals[g] \gets \max(p, \mathbb{P}_\goals[g])$
            \EndFor \label{alg:dtw_end_for}
            \State \Return $\displaystyle \argmax_{\goalstate \in \goals} \mathbb{P}_\goals[\goalstate]$
    \EndWhen
    \end{algorithmic}
\end{algorithm}

\section{On the Projection Function Assumption}

In our main work, we acknowledge a key limitation: either
\begin{enumerate*}[label=(\roman*)]
    \item \abbrev{} has access to the projection function $f$, enabling it to construct a trajectory tree over observations, i.e., $f(\topktrajectoriesnog)$; or
    \item observations and states are assumed to be identical, such that the projection function is the identity, i.e., $f(s)=s$.
\end{enumerate*}
This constraint arises because the trajectory tree path signatures are computed over $\topktrajectoriesnog$, which are sequences of states rather than raw observations.
A promising direction to overcome this limitation is to learn the projection function $f$ directly from data.
In many machine learning contexts, particularly those involving high-dimensional sensory inputs such as images, audio, or text, it is standard to employ representation learning techniques that map observations into a latent space capturing the underlying state dynamics.
This learned representation can then serve as a proxy for the proper projection function, allowing for the construction of trajectory trees based on observations.

Approaches such as autoencoders~\cite{hinton2006reducing}, contrastive learning~\cite{chen2020simple}, and inverse dynamics modeling~\cite{pathak2017curiosity} have demonstrated strong performance in learning such mappings in both supervised and self-supervised settings.
Consequently, while \abbrev{} currently assumes access to $f$ or an identity mapping between states and observations, future work could incorporate a learned projection function to extend the method to more realistic scenarios where the state is not directly observable.

In summary, while our method currently assumes access to a projection function $f$ or an identity mapping between states and observations, this limitation is often not prohibitive in practice.
Many environments (particularly in simulation, control, and structured domains) do not distinguish between states and observations, effectively rendering $f$ trivial.
Moreover, in cases where such a distinction exists, robust and well-established solutions already exist for inferring or approximating the latent state space.
As such, the reliance on $f$ does not significantly constrain the applicability of our approach and can be addressed through either learned or engineered alternatives, depending on the context.

\section{Path Signature Depth Selection}

To investigate the impact of varying path signature depths on the performance of \abbrev{} and \abbrev{}+DTW, we conduct experiments using four different depth values: $\left[2, 3, 4, 5\right]$.
As discussed in Section~\ref{sec:computing_path_signatures}, increasing the depth of a path signature expands the representational space by incorporating higher-order interactions among input features, potentially yielding more informative representations.
However, excessively deep signatures may introduce spurious or non-ideal relationships between state features.
It is also important to note that the computational complexity of both inference modes scales with $d^k$, where $d$ denotes the number of input dimensions and $k$ the signature depth.
Thus, if shallower signatures prove sufficient, \abbrev{} can benefit from improved computational efficiency.

Table~\ref{tab:k_results} displays the results for these experiments.
For \abbrev, both PPV and ACC exhibit a slight downward trend as $k$ increases, with PPV decreasing from $54.1\%$ to $52.8\%$ and ACC from $86.9\%$ to $86.5\%$.
Online Time increases modestly from $0.017$ to $0.021$ seconds, while Offline Time remains relatively stable around $631$ seconds.
Similarly, \abbrev{}+DTW exhibits a slight decline in PPV (from $56.3\%$ to $56.0\%$) and ACC (from $87.5\%$ to $87.4\%$) as depth increases.
However, Online Time grows more substantially, rising from $11.552$ to $15.805$ seconds, whereas Offline Time remains nearly constant, fluctuating slightly around $630$ seconds.
These trends suggest that increasing signature depth introduces marginal performance degradation and higher online computational costs, while offline processing remains largely unaffected.

\begin{table}[h!tbp]
\scriptsize
\centering
\begin{tabular*}{\columnwidth}{ll@{\extracolsep{\fill}}rrrr}
\toprule
Method & Metric             & 2 & 3 & 4 & 5 \\ \midrule
\multirow{4}{*}{\abbrev{}}  & PPV (\%) & 54.1 & 53.9 & 53.4 & 52.8 \\
                            & ACC (\%) & 86.9 & 86.8 & 86.7 & 86.5 \\
                            & Online Time & 0.017 & 0.017 & 0.019 & 0.021 \\
                            & Offline Time & 631.006 & 631.210 & 631.346 & 631.598 \\  \midrule
\multirow{4}{*}{\abbrev{}+DTW}  & PPV (\%) & 56.3 & 56.2 & 56.2 & 56.0 \\
                                & ACC (\%) & 87.5 & 87.5 & 87.5 & 87.4 \\
                                & Online Time & 11.552 & 14.540 & 15.303 & 15.805 \\
                                & Offline Time & 630.313 & 630.439 & 630.538 & 630.838 \\ 
\bottomrule
\end{tabular*}
\caption{Results for different $k$ values}
\label{tab:k_results}
\end{table}

\section{Grid Search Results}

Tables~\ref{tab:GRPS} and~\ref{tab:GRPS+DTW} show the grid search results over 10 scenarios from the \citeauthor{tesch2023online}~\shortcite{tesch2023online} dataset.
The table displays the results with the maximum PPV for each merge and prune variable, along with the respective $\topksize$ value.

\section{Experiment Results}

Tables~\ref{tab:GRPS_all} and~\ref{tab:GRPS_DTW_all} show the experiments results for the whole dataset of \citeauthor{tesch2023online}\shortcite{tesch2023online} using Top-$\topksize$ value of $\topksize=15$. The GRPS method experiments are executed using merge and prune of 0.2 for both variables. The GRPS+DTW method uses the values of 0.2 and 0.6 for merge and prune variables respectively.

\section{Sampler Quality Evaluation}

We conduct an experimental evaluation to analyze how the trajectory sampler quality influences the performance of our goal recognition approach.
Tables~\ref{tab:sample_quality_grps} and~\ref{tab:sample_quality_grps_dtw} show the results for the GRPS and GRPS+DTW approaches, respectively, considering the first ten scenarios. 
We conducted further experiments with a degraded sampling method, running $RRT^*$ with a shorter running limit of $5$ seconds, instead of $20$ seconds used by~\citet{tesch2023online}.
A comparing between Tables~\ref{tab:sample_quality_grps} and~\ref{tab:sample_quality_grps_dtw} with Tables~\ref{tab:GRPS_all} and~\ref{tab:GRPS_DTW_all}, reveal a clear and directly proportional relationship between trajectory sampling quality and goal inference performance: the closer a sampled trajectory lies to the optimal one, the more accurate the resulting goal inference.

\section{Discrete Domains Experiments}

Tables~\ref{tab:discrete_GRPS} and \ref{tab:discrete_GRPS+DTW} show \abbrev and \abbrev{}+DTW results for the validation split, respectively.
Recall, that we use a $5$ problems from each environment to find the best thresholds for merge and prune before applying it to the full set.
In it, we observe that the best values for merge and prune remain zero.
Different than the findings from the main work, where merge and prune values remained approximately $0.6$ and $0.4$, respectively, results for the discrete environments deteriorated with any values greater than $0$.
We hypothesize that this behavior is inherited from the fact that path signatures use geometrical information to semantically encode different trajectories.
Yet, discrete environments states either:
(i) encode information in a non geometrical manner (e.g., one-hot encodings where values represent categorical states without any inherent spatial or geometric relationships), or
(ii) represent transitions that are abrupt and lack continuity, making it difficult for path signatures to capture meaningful patterns.
Consequently, the deterioration in performance with non-zero merge and prune thresholds suggests that the semantic structure captured by path signatures is disrupted when applied to discrete state representations. 
Table~\ref{tab:discrete_both_all} display the results for \abbrev{} and \abbrev{}+DTW to the whole problem set using merge and prune thresholds $0.0$.
Notably, the performance reported in Table~\ref{tab:discrete_both_all} remains competitive with other goal recognition approaches in discrete settings (when compared to results from \citet{tesch2023online}), underscoring the robustness and versatility of \abbrev{}.

\begin{table*}[ht!]
\scriptsize
\centering
\setlength{\tabcolsep}{4pt}
\begin{tabular*}{\textwidth}{l @{\extracolsep{\fill}} rrrrrrrrrrrr}
\toprule
 & \multicolumn{12}{c}{GRPS} \\ \cmidrule{2-13} 
 & \multicolumn{3}{c}{K=1} & \multicolumn{3}{c}{K=5} & \multicolumn{3}{c}{K=10} & \multicolumn{3}{c}{K=15} \\ \cmidrule{2-4} \cmidrule{5-7} \cmidrule{8-10} \cmidrule{11-13}
Maps & PPV (\%) & Merge & {Prune} & PPV (\%) & Merge & {Prune} & PPV (\%) & Merge & {Prune} & PPV (\%) & Merge & Prune \\ \cmidrule{1-1} \cmidrule{2-4} \cmidrule{5-7} \cmidrule{8-10} \cmidrule{11-13}
{Aftershock}        & 59.2 & 0 & {0.0} & 65.8 & 0.00 & {0.0} & 72.2 & 1.0 & {0.0} & 63.4 & 0.0 & 0.0 \\
{Archipelago}       & 55.7 & 0 & {0.0} & 57.1 & 0.40 & {0.6} & 59.2 & 0.4 & {0.0} & 59.2 & 0.2 & 0.2 \\
{BigGameHunters}    & 53.0 & 0 & {1.0} & 59.2 & 0.60 & {0.0} & 58.3 & 0.2 & {0.2} & 58.0 & 1.2 & 1.2 \\
{Caldera}           & 64.6 & 0 & {0.2} & 69.6 & 0.20 & {0.4} & 66.7 & 2.0 & {0.4} & 65.5 & 1.6 & 1.6 \\
{Circle}            & 80.1 & 0 & {0.4} & 88.1 & 2.00 & {0.0} & 87.2 & 0.2 & {0.0} & 86.3 & 0.2 & 0.0 \\
{CrescentMoon}      & 52.1 & 0 & {0.2} & 58.0 & 0.60 & {0.4} & 54.8 & 0.0 & {0.0} & 56.8 & 0.6 & 0.0 \\
{Desolation}        & 56.3 & 0 & {0.4} & 62.2 & 0.00 & {0.2} & 60.7 & 0.8 & {0.0} & 62.5 & 2.0 & 0.2 \\
{EbonLakes}         & 53.6 & 0 & {0.2} & 57.4 & 1.20 & {0.4} & 54.8 & 0.6 & {0.2} & 56.5 & 0.8 & 0.2 \\
{Entanglement}      & 62.8 & 0 & {0.0} & 66.4 & 0.40 & {0.2} & 65.8 & 0.4 & {0.0} & 64.6 & 0.4 & 0.2 \\
{Eruption}          & 49.7 & 0 & {0.4} & 55.1 & 0.20 & {2.0} & 52.7 & 0.4 & {1.4} & 53.3 & 0.6 & 2.0 \\ \cmidrule{1-1} \cmidrule{2-4} \cmidrule{5-7} \cmidrule{8-10} \cmidrule{11-13}
{Average}           & 58.7 & 0 & {0.3} & 63.9 & 0.56 & {0.4} & 63.2 & 0.6 & {0.2} & 62.6 & 0.8 & 0.6 \\ \bottomrule
\end{tabular*}
\caption{GRPS grid search results for 10 scenarios on the dataset.}
\label{tab:GRPS}
\end{table*}

\begin{table*}[ht!]
\scriptsize
\centering
\setlength{\tabcolsep}{4pt}
\begin{tabular*}{\textwidth}{l @{\extracolsep{\fill}} rrrrrrrrrrrr}
\toprule
 & \multicolumn{12}{c}{GRPS+DTW} \\ \cmidrule{2-13} 
 & \multicolumn{3}{c}{K=1} & \multicolumn{3}{c}{K=5} & \multicolumn{3}{c}{K=10} & \multicolumn{3}{c}{K=15} \\ \cmidrule{2-4} \cmidrule{5-7} \cmidrule{8-10} \cmidrule{11-13}
Maps & PPV (\%) & Merge & {Prune} & PPV (\%) & Merge & {Prune} & PPV (\%) & Merge & {Prune} & PPV (\%) & Merge & Prune \\ \cmidrule{1-1} \cmidrule{2-4} \cmidrule{5-7} \cmidrule{8-10} \cmidrule{11-13}
{Aftershock}    & 62.5 & 0.0 & {0.6} & 64.6 & 1.0 & {0.6} & 65.8 & 0.0 & {0.0} & 65.5 & 1.0 & 0.0 \\
{Archipelago}   & 55.1 & 0.0 & {0.0} & 58.9 & 0.6 & {0.2} & 57.7 & 2.0 & {0.0} & 57.4 & 1.0 & 0.6 \\
{BigGameHunters}& 58.6 & 0.0 & {2.0} & 61.0 & 0.0 & {2.0} & 61.3 & 1.6 & {1.6} & 61.6 & 0.0 & 1.8 \\
{Caldera}       & 67.9 & 0.0 & {0.8} & 71.4 & 0.0 & {1.0} & 72.6 & 0.0 & {0.8} & 72.6 & 0.0 & 0.8 \\
{Circle}        & 79.8 & 0.0 & {0.6} & 88.4 & 1.8 & {0.0} & 88.4 & 0.0 & {0.6} & 88.4 & 0.4 & 0.0 \\
{CrescentMoon}  & 53.6 & 0.0 & {1.4} & 57.4 & 0.2 & {1.4} & 58.3 & 1.2 & {0.6} & 56.8 & 0.0 & 1.4 \\
{Desolation}    & 57.7 & 0.0 & {0.6} & 62.8 & 1.2 & {0.0} & 62.5 & 1.8 & {0.8} & 63.1 & 0.4 & 0.8 \\
{EbonLakes}     & 54.5 & 0.0 & {0.8} & 60.4 & 0.8 & {0.6} & 58.6 & 2.0 & {0.0} & 56.8 & 0.6 & 0.2 \\
{Entanglement}  & 67.3 & 0.0 & {0.6} & 68.5 & 0.2 & {0.6} & 68.5 & 0.0 & {0.6} & 67.9 & 2.0 & 0.0 \\
{Eruption}      & 52.1 & 1.8 & {0.0} & 55.7 & 1.4 & {0.8} & 56.0 & 0.2 & {0.6} & 53.9 & 1.6 & 0.6 \\ \cmidrule{1-1} \cmidrule{2-4} \cmidrule{5-7} \cmidrule{8-10} \cmidrule{11-13}
{Average}       & 60.9 & 0.2 & {0.7} & 64.9 & 0.7 & {0.7} & 65.0 & 0.9 & {0.6} & 64.4 & 0.6 & 0.6 \\ \bottomrule
\end{tabular*}
\caption{GRPS+DTW grid search results for 10 scenarios on the dataset.}
\label{tab:GRPS+DTW}
\end{table*}

\begin{table*}[!htbp]
\scriptsize
\centering
\begin{tabular*}{.8\textwidth}{l @{\extracolsep{\fill}} rrrrrr}
\toprule
 & \multicolumn{6}{c}{GRPS} \\ \cmidrule{2-7} 
Maps & PPV (\%) & ACC (\%) & SPR & PC & Online Time (s) & Offline Time(s) \\ \midrule
{Aftershock}        & 54.8	& 87.0	& 1.0	& 7.0& 0.023& 267.9 \\
{Archipelago}       & 50.0	& 85.7	& 1.0	& 7.0& 0.021
& 385.0 \\
{BigGameHunters}    & 44.6	& 84.1	& 1.0	& 7.0& 0.025
& 510.2 \\
{Caldera}           & 53.3	& 86.6	& 1.0	& 7.0& 0.025
& 504.5 \\
{Circle}            & 73.2	& 92.3	& 1.0	& 7.0& 0.020
& 289.6 \\
{CrescentMoon}      & 45.8	& 84.5	& 1.0	& 7.0& 0.024
& 332.3 \\
{Desolation}        & 48.8	& 85.4	& 1.0	& 7.0& 0.017
& 258.2 \\
{EbonLakes}         & 35.9	& 81.3	& 1.1	& 7.0& 0.017
& 325.6 \\
{Entanglement}      & 44.6	& 84.2	& 1.0	& 7.0& 0.016
& 273.5 \\
{Eruption}          & 40.2	& 82.6	& 1.1	& 7.0& 0.016
& 185.8 \\
\bottomrule
\end{tabular*}
\caption{GRPS method experiments results for sampler quality evaluation.}
\label{tab:sample_quality_grps}
\end{table*}

\begin{table*}[!htbp]
\scriptsize
\centering
\begin{tabular*}{.8\textwidth}{l @{\extracolsep{\fill}} rrrrrr}
\toprule
 & \multicolumn{6}{c}{GRPS} \\ \cmidrule{2-7} 
Maps & PPV (\%) & ACC (\%) & SPR & PC & Online Time (s) & Offline Time(s) \\ \midrule
{Aftershock}        & 54.5	& 87.0	& 1.0	& 7.0	& 3.5	& 248.1\\
{Archipelago}       & 44.6	& 84.2	& 1.0	& 7.0	& 5.8	& 366.5\\
{BigGameHunters}    & 46.1	& 84.6	& 1.0	& 7.0	& 10.8	& 537.3\\
{Caldera}           & 60.1	& 88.6	& 1.0	& 7.0	& 6.2	& 496.3\\
{Circle}            & 75.9	& 93.1	& 1.0	& 7.0	& 3.4	& 238.7\\
{CrescentMoon}      & 48.5	& 85.3	& 1.0	& 7.0	& 5.5	& 345.9\\
{Desolation}        & 54.2	& 86.9	& 1.0	& 7.0	& 3.7	& 268.9\\
{EbonLakes}         & 44.0	& 83.8	& 1.1	& 7.0	& 5.4	& 329.2\\
{Entanglement}      & 45.2	& 84.4	& 1.0	& 7.0	& 4.0	& 332.1\\
{Eruption}          & 45.6	& 84.4	& 1.0	& 7.0	& 3.7	& 194.4\\
\bottomrule
\end{tabular*}
\caption{GRPS+DTW method experiments results for sampler quality evaluation.}
\label{tab:sample_quality_grps_dtw}
\end{table*}

\begin{table*}[!htbp]
\scriptsize
\centering
\begin{tabular*}{.8\textwidth}{l @{\extracolsep{\fill}} rrrrrr}
\toprule
 & \multicolumn{6}{c}{GRPS} \\ \cmidrule{2-7} 
Maps & PPV (\%) & ACC (\%) & SPR & PC & Online Time (s) & Offline Time(s) \\ \midrule
{Aftershock}        & 63.1 & 89.5 & 1 & 7 & 0.029 & 310 \\
{Archipelago}       & 56.8 & 87.7 & 1 & 7 & 0.028 & 300 \\
{BigGameHunters}    & 54.5 & 87.0 & 1 & 7 & 0.032 & 390 \\
{Caldera}           & 68.8 & 91.1 & 1 & 7 & 0.030 & 340 \\
{Circle}            & 85.7 & 95.9 & 1 & 7 & 0.029 & 260 \\
{CrescentMoon}      & 54.8 & 87.1 & 1 & 7 & 0.029 & 310 \\
{Desolation}        & 56.5 & 87.6 & 1 & 7 & 0.027 & 250 \\
{EbonLakes}         & 50.6 & 85.9 & 1 & 7 & 0.029 & 310 \\
{Entanglement}      & 64.0 & 89.7 & 1 & 7 & 0.026 & 220 \\
{Eruption}          & 48.8 & 85.4 & 1 & 7 & 0.031 & 300 \\
{HotZone}           & 61.9 & 89.1 & 1 & 7 & 0.031 & 300 \\
{Isolation}         & 61.3 & 88.9 & 1 & 7 & 0.029 & 320 \\
{Legacy}            & 36.0 & 81.7 & 1 & 7 & 0.028 & 230 \\
{OrbitalGully}      & 54.8 & 87.1 & 1 & 7 & 0.029 & 320 \\
{Predators}         & 59.8 & 88.5 & 1 & 7 & 0.027 & 230 \\
{Ramparts}          & 43.5 & 83.8 & 1 & 7 & 0.031 & 430 \\
{RedCanyons}        & 58.0 & 88.0 & 1 & 7 & 0.028 & 290 \\
{Rosewood}          & 50.6 & 85.9 & 1 & 7 & 0.031 & 340 \\
{Sanctuary}         & 44.0 & 84.0 & 1 & 7 & 0.033 & 400 \\
{ShroudPlatform}    & 45.8 & 84.5 & 1 & 7 & 0.034 & 370 \\
{SpaceAtoll}        & 38.7 & 82.5 & 1 & 7 & 0.030 & 340 \\
{SpaceDebris}       & 45.8 & 84.5 & 1 & 7 & 0.031 & 330 \\
{SteppingStones}    & 73.8 & 92.5 & 1 & 7 & 0.027 & 200 \\
{Triskelion}        & 48.8 & 85.4 & 1 & 7 & 0.032 & 390 \\
{WarpGates}         & 48.5 & 85.3 & 1 & 7 & 0.028 & 320 \\
{WatersEdge}        & 43.7 & 83.9 & 1 & 7 & 0.032 & 450 \\
{WaypointJunction}  & 51.5 & 86.1 & 1 & 7 & 0.028 & 250 \\
{WinterConquest}    & 46.7 & 84.8 & 1 & 7 & 0.032 & 380 \\ \midrule
{Average}           & 54.2 & 86.9 & 1 & 7 & 0.030 & 320 \\ \bottomrule
\end{tabular*}
\caption{GRPS method experiments results for the whole scenarios on the dataset.}
\label{tab:GRPS_all}
\end{table*}

\begin{table*}[!htbp]
\scriptsize
\centering
\begin{tabular*}{.8\textwidth}{l  @{\extracolsep{\fill}} rrrrrr}
\toprule
 & \multicolumn{6}{c}{GRPS+DTW} \\ \cmidrule{2-7} 
Maps & PPV (\%) & ACC (\%) & SPR & PC & Online Time (s) & Offline Time (s) \\ \midrule
{Aftershock}        & 60.7 & 88.8 & 1 & 7 & 18.0 & 260 \\
{Archipelago}       & 56.0 & 87.4 & 1 & 7 & 18.0 & 260 \\
{BigGameHunters}    & 59.5 & 88.4 & 1 & 7 & 25.0 & 330 \\
{Caldera}           & 65.8 & 90.2 & 1 & 7 & 21.0 & 280 \\
{Circle}            & 84.8 & 95.7 & 1 & 7 & 13.0 & 220 \\
{CrescentMoon}      & 56.3 & 87.5 & 1 & 7 & 18.0 & 260 \\
{Desolation}        & 61.9 & 89.1 & 1 & 7 & 12.0 & 220 \\
{EbonLakes}         & 53.0 & 86.6 & 1 & 7 & 19.0 & 260 \\
{Entanglement}      & 61.9 & 89.1 & 1 & 7 & 10.0 & 200 \\
{Eruption}          & 53.9 & 86.8 & 1 & 7 & 18.0 & 250 \\
{HotZone}           & 57.4 & 87.8 & 1 & 7 & 17.0 & 250 \\
{Isolation}         & 63.4 & 89.5 & 1 & 7 & 18.0 & 270 \\
{Legacy}            & 43.5 & 83.8 & 1 & 7 & 12.0 & 200 \\
{OrbitalGully}      & 58.9 & 88.3 & 1 & 7 & 19.0 & 270 \\
{Predators}         & 61.0 & 88.9 & 1 & 7 & 11.0 & 200 \\
{Ramparts}          & 43.8 & 83.9 & 1 & 7 & 30.0 & 350 \\
{RedCanyons}        & 60.1 & 88.6 & 1 & 7 & 16.0 & 250 \\
{Rosewood}          & 53.6 & 86.7 & 1 & 7 & 22.0 & 290 \\
{Sanctuary}         & 45.5 & 84.4 & 1 & 7 & 26.0 & 340 \\
{ShroudPlatform}    & 49.7 & 85.6 & 1 & 7 & 28.0 & 320 \\
{SpaceAtoll}        & 42.9 & 83.7 & 1 & 7 & 25.0 & 310 \\
{SpaceDebris}       & 45.8 & 84.5 & 1 & 7 & 22.0 & 280 \\
{SteppingStones}    & 76.2 & 93.2 & 1 & 7 &  8.3 & 170 \\
{Triskelion}        & 52.7 & 86.5 & 1 & 7 & 25.0 & 310 \\
{WarpGates}         & 53.3 & 86.6 & 1 & 7 & 19.0 & 270 \\
{WatersEdge}        & 49.4 & 85.5 & 1 & 7 & 30.0 & 370 \\
{WaypointJunction}  & 53.6 & 86.7 & 1 & 7 & 14.0 & 230 \\
{WinterConquest}    & 54.2 & 86.9 & 1 & 7 & 27.0 & 330 \\ \midrule
{Average}           & 56.4 & 87.5 & 1 & 7 & 19.0 & 270 \\
\bottomrule
\end{tabular*}
\caption{GRPS+DTW method experiments results for the whole scenarios on the dataset.}
\label{tab:GRPS_DTW_all}
\end{table*}

\begin{table*}[!htbp]
\scriptsize
\centering
\setlength{\tabcolsep}{4pt}
\begin{tabular*}{\textwidth}{l @{\extracolsep{\fill}} rrr rrr rrr rrr}
\toprule
 & \multicolumn{12}{c}{GRPS} \\ \cline{2-13} 
 & \multicolumn{3}{c}{K=1} & \multicolumn{3}{c}{K=5} & \multicolumn{3}{c}{K=10} & \multicolumn{3}{c}{K=15} \\ \cmidrule{2-4} \cmidrule{5-7} \cmidrule{8-10} \cmidrule{11-13}
Scenario & PPV (\%) & Merge & {Prune} & PPV (\%) & Merge & {Prune} & PPV (\%) & Merge & {Prune} & PPV (\%) & Merge & Prune \\ \cmidrule{1-1} \cmidrule{2-4} \cmidrule{5-7} \cmidrule{8-10} \cmidrule{11-13}
{blocks-world}  & 5.0 & 0.0 & {0.0}  & 36.0 & 0.0 & {0.0}   & 38.4 & 0.0 & {0.0} & 39.0 & 0.0 & 0.0 \\
{depots}        & 55.7 & 0.0 & {0.0}  & 43.4 & 0.0 & {0.0}   & 43.9 & 0.0 & {0.0} & 41.0 & 0.0 & 0.0 \\
{driverlog}     & 15.3 & 0.0 & {0.0}  & 38.4 & 0.0 & {0.0}   & 45.9 & 0.0 & {0.0} & 46.0 & 0.0 & 0.0 \\
{dwr}           & 15.3 & 0.0 & {0.0}  & 40.8 & 0.0 & {0.0}   & 39.8 & 0.0 & {0.0} & 40.3 & 0.0 & 0.0 \\
{ferry}         & 16.4 & 0.0 & {0.0}  & 57.8 & 0.0 & {0.0}   & 65.7 & 0.0 & {0.0} & 70.4 & 0.0 & 0.0 \\ \cmidrule{1-1} \cmidrule{2-4} \cmidrule{5-7} \cmidrule{8-10} \cmidrule{11-13}
{Average}       & 12.9 & 0.0 & {0.0}  & 43.3 & 0.0 & {0.0}   & 46.7 & 0.0 & {0.0} & 47.3 & 0.0 & 0.0 \\
\bottomrule
\end{tabular*}
\caption{GRPS grid search results for 5 scenarios on the discrete domain dataset.}
\label{tab:discrete_GRPS}
\end{table*}

\begin{table*}[!htbp]
\scriptsize
\centering
\setlength{\tabcolsep}{4pt}
\begin{tabular*}{\textwidth}{l @{\extracolsep{\fill}} rrr rrr rrr rrr}
\toprule
 & \multicolumn{12}{c}{GRPS+DTW} \\ \cline{2-13} 
 & \multicolumn{3}{c}{K=1} & \multicolumn{3}{c}{K=5} & \multicolumn{3}{c}{K=10} & \multicolumn{3}{c}{K=15} \\ \cmidrule{2-4} \cmidrule{5-7} \cmidrule{8-10} \cmidrule{11-13}
Scenario & PPV (\%) & Merge & {Prune} & PPV (\%) & Merge & {Prune} & PPV (\%) & Merge & {Prune} & PPV (\%) & Merge & Prune \\ \cmidrule{1-1} \cmidrule{2-4} \cmidrule{5-7} \cmidrule{8-10} \cmidrule{11-13}
{blocks-world}  & 5.0 & 0.0 & {0.0}  & 36.0 & 0.0 & {0.0}   & 38.4 & 0.0 & {0.0} & 39.0 & 0.0 & 0.0 \\
{depots}        & 55.7 & 0.0 & {0.0}  & 43.4 & 0.0 & {0.0}   & 43.9 & 0.0 & {0.0} & 41.0 & 0.0 & 0.0 \\
{driverlog}     & 15.3 & 0.0 & {0.0}  & 38.4 & 0.0 & {0.0}   & 45.9 & 0.0 & {0.0} & 46.0 & 0.0 & 0.0 \\
{dwr}           & 15.3 & 0.0 & {0.0}  & 40.8 & 0.0 & {0.0}   & 39.8 & 0.0 & {0.0} & 40.3 & 0.0 & 0.0 \\
{ferry}         & 16.4 & 0.0 & {0.0}  & 59.0 & 0.0 & {0.0}   & 66.4 & 0.0 & {0.0} & 69.8 & 0.0 & 0.0 \\ \cmidrule{1-1} \cmidrule{2-4} \cmidrule{5-7} \cmidrule{8-10} \cmidrule{11-13}
{Average}       & 12.9 & 0.0 & {0.0}  & 43.5 & 0.0 & {0.0}   & 46.9 & 0.0 & {0.0} & 47.2 & 0.0 & 0.0 \\
\bottomrule
\end{tabular*}
\caption{GRPS+DTW grid search results for 5 scenarios on the discrete domain dataset.}
\label{tab:discrete_GRPS+DTW}
\end{table*}

\begin{table*}[!htbp]
\scriptsize
\centering
\begin{tabular*}{\textwidth}{l @{\extracolsep{\fill}} rrrrrrrr rrrrrr}
\toprule
 &  &  & \multicolumn{6}{c}{GRPS K=15} & \multicolumn{6}{c}{GRPS+DTW K=15} \\ \cmidrule(r){4-9} \cmidrule(l){10-15}
Scenario       & $\vert O \vert$ & $\vert G \vert$ & PPV & Acc & Spread & PC & Online & Offline & PPV & Acc & Spread & PC & Online & Offline \\ \cmidrule(r){1-3} \cmidrule(r){4-9} \cmidrule(l){10-15}
blocks-world & 8.8 & 20.0 & 39.0 & 91.1 & 2.1 & 20.0 & 1.5 & 49.5 & 39.0 & 91.1	&2.1	&20.0	&11.5	&52.3\\
depots & 9.5 & 8.0 & 41.0 & 81.3 & 2.0 & 8.0 & 0.3 & 12.6 & 41.0	& 81.3	&1.9	&8.0	&2.9	&14.1\\
driverlog & 12.2 & 6.7 & 46.0 & 83.4 & 1.2 & 6.7 & 0.2 & 11.5 & 46.0	&83.4	&1.2	&6.6	&2.4	&12.5\\
dwr & 22.0 & 6.7 & 40.3 & 77.4 & 2.2 & 6.7 & 7.1 & 20.9 & 40.3 &77.4	&2.2	&6.7 &122.7	&21.6\\
easy-ipc-grid & 11.9 & 7.5 & 61.6 & 89.4 & 1.3 & 7.5 & 13.9 & 52.0 & 61.6	&89.4	&1.3 &7.5	&107.6	&39.8\\
ferry & 18.8 & 6.3 & 70.4 & 90.4 & 1.0 & 6.3 & 0.2 & 7.4 & 69.8	&90.3 &1.0	&6.3	&4.2	&7.9\\
logistics & 18.2 & 10.0 & 61.2 & 92.2 & 1.3 & 10.0 & 1.2 & 19.1 & 61.2	&92.2	&1.3	&10.0	&19.1	&18.7\\
miconic & 16.3 & 6.0 & 61.6 & 87.3 & 1.2 & 6.0 & 0.1 & 10.2 & 61.6	& 87.3	&1.2	&6.0	&2.5	&10.2\\
rovers & 10.8 & 6.0 & 53.0 & 83.4 & 1.4 & 6.0 & 0.2 & 211.2 & 53.0	&83.4	&1.4	&6.0	&2.6	&227.6\\
satellite & 10.8 & 6.0 & 41.0 & 72.7 & 2.4 & 6.0 & 0.1 & 8.6 & 41.0	&72.7 &2.4	&6.0	&1.5	&10.3\\
small-sokoban & 26.0 & 9.0 & 53.7 & 90.2 & 1.6 & 9.0 & 96.1 & 204.2 & 53.7	&90.2	&1.6	&9.0	&891.9	&207.2\\
zeno-travel & 12.0 & 6.0 & 55.2 & 85.0 & 1.2 & 6.0 & 0.2 & 132.2 & 55.2	&85.0	&1.2	&6.0	&1.7	&132.7\\ 
\cmidrule(r){1-3} \cmidrule(r){4-9} \cmidrule(l){10-15}
Average        & 14.8 & 8.2 & 52.0 & 85.3 & 1.6 & 8.2 & 10.1 & 60.6 & 51.9 & 85.3 & 1.6 & 8.2 & 97.6 & 62.9 \\
\bottomrule
\end{tabular*}
\caption{Discrete domains experiments results of goal recognition using path signature method.}
\label{tab:discrete_both_all}
\end{table*}

\end{document}